\documentclass[letterpaper, 10 pt, conference]{ieeeconf}  
\usepackage[pdftex]{graphicx}

\usepackage{subcaption}
\usepackage{booktabs}
\usepackage{multirow}
\usepackage{pifont}
\usepackage{newtxtext,newtxmath}

\usepackage{caption}
\DeclareCaptionFont{eightpt}{\fontsize{8pt}{9.6pt}\selectfont}

\IEEEoverridecommandlockouts                              

\usepackage{url}

\usepackage{graphics} 

\title{\LARGE \bf
Uncertainty-Aware Autonomous Vehicles: Predicting the Road Ahead}

\author{
  \begin{tabular}{c}
    \textbf{Shireen Kudukkil Manchingal}\textsuperscript{$1\dagger$}\thanks{$\dagger$ Corresponding author: \textit{smanchingal@brookes.ac.uk}} \ \ \ \ \ \ \
    \textbf{Armand Amaritei}\textsuperscript{$2$}\ \ \ \ \ \ \
    \textbf{Mihir Gohad}\textsuperscript{$2$}\ \ \ \ \ \ \
    \textbf{Maryam Sultana}\textsuperscript{$1$} \\
    \textbf{Julian F. P. Kooij}\textsuperscript{$3$}\ \ \ \ \ \ \
    \textbf{Fabio Cuzzolin}\textsuperscript{$1$}\ \ \ \ \ \ \
    \textbf{Andrew Bradley}\textsuperscript{$1,2$} \\
    \textsuperscript{$1$}School of Engineering, Computing and Mathematics, Oxford Brookes University, UK\\
    \textsuperscript{$2$}Autonomous Driving and Intelligent Transport, Oxford Brookes University, UK\\
    \textsuperscript{$3$}Cognitive Robotics, TU Delft, Netherlands\\
    \thanks{* This work has received funding from the European Union’s Horizon 2020 Research and Innovation program under Grant Agreement No. 964505 (E-pi).}
  \end{tabular}
}

\begin{document}

\maketitle
\thispagestyle{empty}
\pagestyle{empty}

\begin{abstract}

Autonomous Vehicle (AV) perception systems have advanced rapidly in recent years, providing vehicles with the ability to accurately interpret their environment. Perception systems remain susceptible to errors caused by overly-confident predictions in the case of rare events or out-of-sample data. This study equips an autonomous vehicle with the ability to 'know when it is uncertain', using an uncertainty-aware image classifier as part of the AV software stack. Specifically, the study exploits the ability of Random-Set Neural Networks (RS-NNs) to explicitly quantify prediction uncertainty. Unlike traditional CNNs or Bayesian methods, RS-NNs predict belief functions over sets of classes, allowing the system to identify and signal uncertainty clearly in novel or ambiguous scenarios. The system is tested in a real-world autonomous racing vehicle software stack, with the RS-NN classifying the layout of the road ahead and providing the associated uncertainty of the prediction. Performance of the RS-NN under a range of road conditions is compared against traditional CNN and Bayesian neural networks, with the RS-NN achieving significantly higher accuracy and superior uncertainty calibration. This integration of RS-NNs into Robot Operating System (ROS)-based vehicle control pipeline demonstrates that predictive uncertainty can dynamically modulate vehicle speed, maintaining high-speed performance under confident predictions while proactively improving safety through speed reductions in uncertain scenarios. These results demonstrate the potential of uncertainty-aware neural networks - in particular RS-NNs - as a practical solution for safer and more robust autonomous driving. 

\end{abstract}

\section{INTRODUCTION}

Autonomous vehicle (AV) technology has made rapid strides, driven largely by advances in machine perception, \textit{i.e.}, the ability of a vehicle to `see' and interpret its environment. However, despite these gains, perception systems remain a critical bottleneck to AV safety. High-profile failures \cite{boudette2021tesla} have revealed that even state-of-the-art models can falter in `edge cases': rare or out-of-distribution scenarios for which the system has no direct training experience \cite{fursa2021worsening}.
A core issue is that most neural networks and modern perception systems are significantly over-confident in their predictions \cite{ovadia2019can, minderer2021revisiting}, often providing a seemingly certain detection or classification even when there is substantial underlying uncertainty. This overconfidence is at odds with human driving behavior; a human will acknowledge that they are uncertain about the road ahead, and likely slow down, whereas a neural network typically offers a single `best guess', with potentially life-threatening consequences.

The rise of autonomous racing competitions \cite{nguyen2006natcar, motevalli2011application, case1996formula}, emerged a critical testbed for advanced perception, planning, decision-making, and control algorithms to their limits; driven by stringent performance demands and safety constraints \cite{wischnewski2022indy, hullermeier2021aleatoric}, running on low-powered, lightweight hardware that must be inside the vehicle. High-speed autonomous vehicles (AVs) in racing scenarios must execute precise real-time predictions about their immediate environment while ensuring safety under rapidly changing conditions \cite{kabzan2020amz, 10254777}. 

Despite substantial research on robustness, both road-going and racing AVs remain susceptible to failures in the face of rare or unexpected events, such as novel weather patterns or unforeseen track layouts. These scenarios often go undetected as `risky' by current systems, leading to overconfident actions at precisely the moments when caution is most warranted. In racing, this can result in an `off track' event; on the road, it may have even graver consequences.

A key opportunity lies in equipping AVs with the ability to `know when they do not know': to react with caution when uncertainty is high, or to push the performance envelope when it is appropriate to do so. Uncertainty-aware approaches to perception are an emerging area of research aimed at addressing exactly this need.

\begin{figure*}[!ht]
    \centering
    \hspace{-0.4em}
        \includegraphics[width=0.7\textwidth]{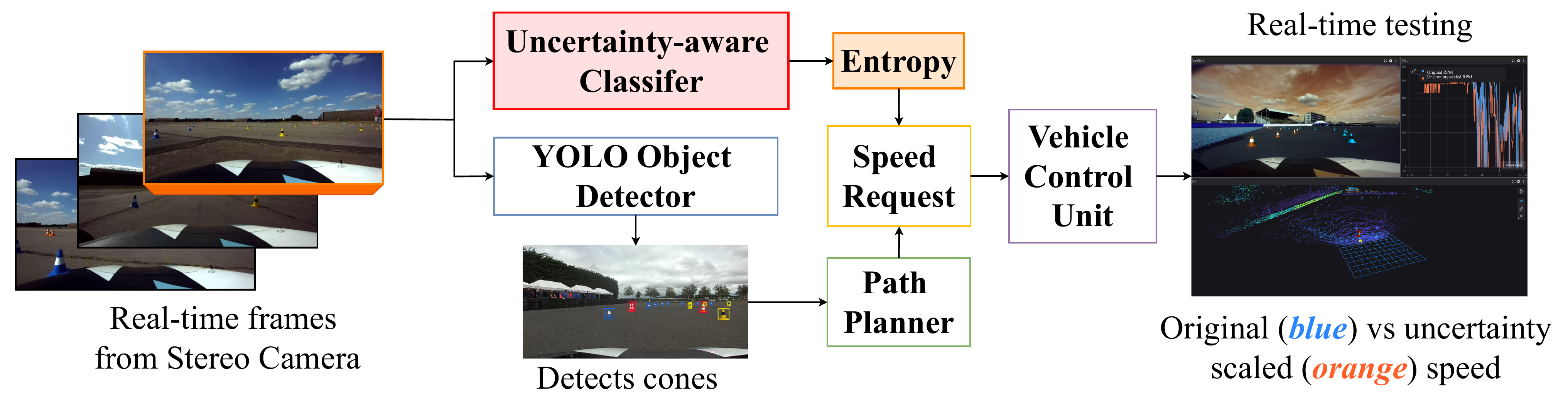} 
  \caption{Real-time uncertainty awareness in the autonomous racing car.}
  \label{fig:ROS}
\end{figure*}

Traditional approaches to uncertainty quantification in computer vision typically rely on Bayesian neural networks \cite{gal2016dropout, hobbhahn2022fast} or ensemble methods \cite{lakshminarayanan2017simple}. However, these approaches often struggle under severe domain shifts or when confronted with limited or ambiguous training data \cite{guo2017calibration}. Recent advances have proposed explicitly modeling uncertainty via Random-Set Neural Networks (RS-NNs) \cite{manchingal2025random}, which predict belief functions over sets of possible outcomes, rather than a single probability distribution. RS-NNs circumvent the need to specify priors, as required by Bayesian methods, and do not require large training times, as in ensembles. They are also important in applications where the data is not too distinct as set relations can learn such nuances. 
Whilst promising approaches, to date such techniques have seen little application in the AV domain. In this work, we aim to fill this gap by evaluating uncertainty-aware models and integrating the most suitable approach into the AV software stack as shown in Fig. \ref{fig:ROS} (detailed in Sec. \ref{sec:uq-av}), thus enabling real-time identification of `unexpected events', in this case, a previously unforeseen road layout - allowing the vehicle to proactively reduce speed in the presence of uncertainty. To the best of our knowledge, this represents a novel application of uncertainty-aware modeling in autonomous driving, directly embedded into the decision-making pipeline.

\section{RELATED WORK} \label{sec:related}

Recent autonomous racing systems have pushed perception and planning to their limits, with teams like TUM and ETH Zurich demonstrating high-speed performance using CNN (Convolutional Neural Network) based cone detection, trajectory optimization, and minimum-curvature path planners \cite{betz2023tum,kabzan2020amz, garlick2022real,andresen2020accurate}. However, these pipelines remain highly sensitive to perception errors, motivating research into robustness through data augmentation and retraining \cite{teeti2022vision,mușat2021multi,fursa2021worsening, 10254777}. However, such methods lack explicit uncertainty quantification, leaving systems vulnerable in ambiguous or unseen scenarios.

Uncertainty quantification (UQ) has therefore emerged as a key factor in reliable AV perception, spanning segmentation \cite{kendall2015bayesian,zhao2022pyramid,sirohi2022uncertainty}, object detection \cite{choi2018uncertainty,oksuz2023towards,feng2021labels}, and depth estimation \cite{gustafsson2020evaluating,qu2021bayesian}. Techniques such as Bayesian SegNet \cite{kendall2015bayesian}, MC dropout \cite{gal2016dropout}, evidential learning \cite{sensoy}, and ensemble \cite{lakshminarayanan2017simple} methods have improved robustness under noise, occlusion, and OOD conditions. More recently, UQ has been extended to end-to-end controllers, enabling real-time steering angle uncertainty estimation \cite{choi2018uncertainty}. Despite progress, major challenges hinder the full potential of UQ in autonomous vehicles. The field lacks ground-truth annotations, standardized evaluation criteria, and robust benchmarks, making it difficult to compare methods or assess real-world impact \cite{wang2025uncertainty}. Many approaches also fail in open-world, out-of-distribution (OOD) scenarios. Most critically, UQ is rarely integrated into control or planning, serving mainly as a diagnostic metric. 

Our work bridges this gap by posing next-turn prediction as a classification problem to enable quick, real-time results, and by evaluating uncertainty-aware models on this task. We integrate the best-performing model directly into an autonomous racing vehicle’s control pipeline, using its uncertainty estimates to dynamically modulate vehicle speed and enhance safety, robustness, and performance under realistic racing conditions. To the best of our knowledge, this is also the first classifier to predict the road layout ahead while providing calibrated uncertainty estimates, enabling proactive and uncertainty-aware planning.

\section{IDENTIFICATION OF ROAD LAYOUT UNDER UNCERTAINTY}
\label{sec:identify-road}

\subsection{The Autonomous Racing Challenge} \label{sec:fs-ai}

\begin{figure}[thpb]
  \centering
  \vspace{4pt}
\includegraphics[width=0.36\textwidth]{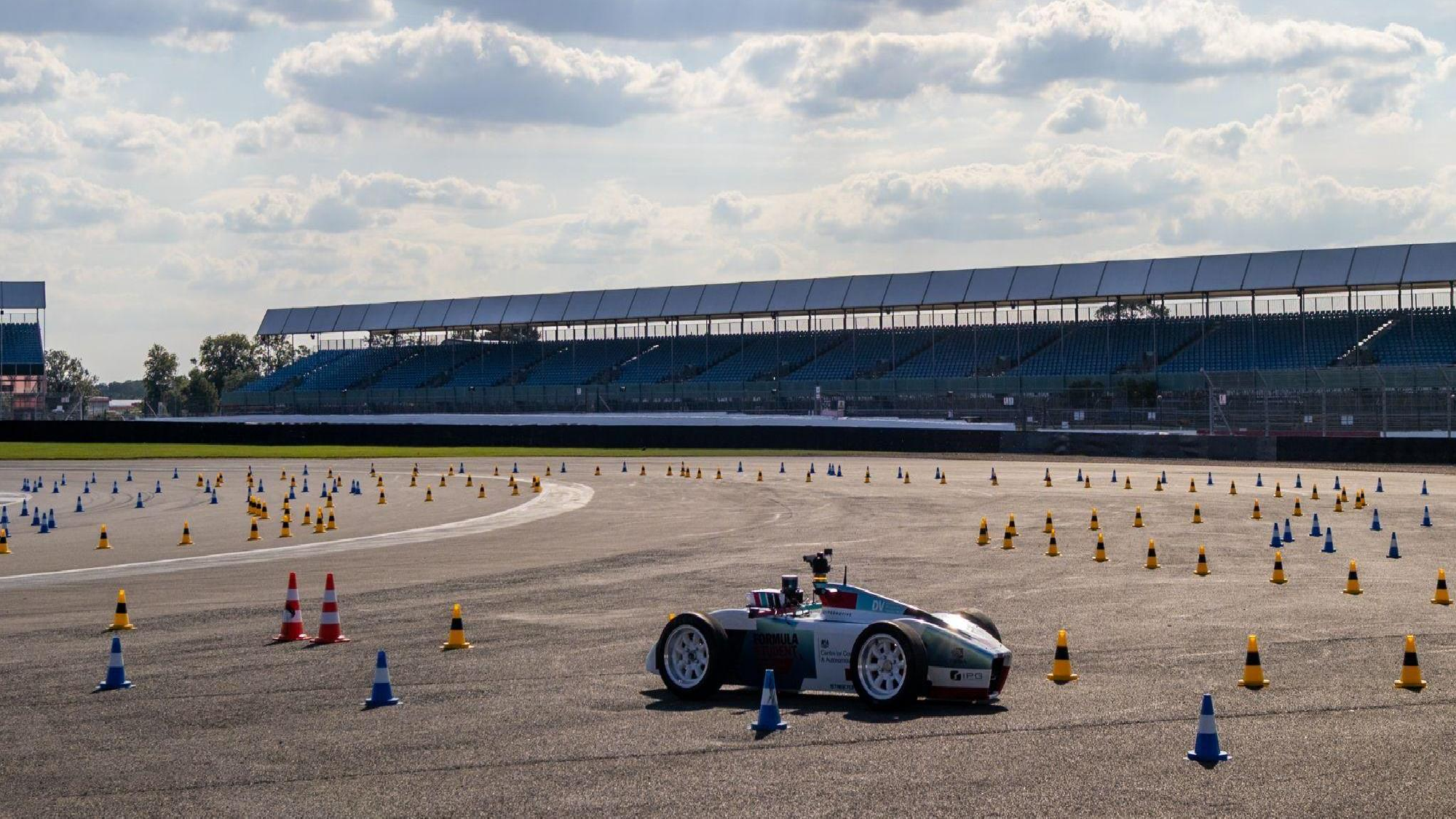} 
  \caption{The IMechE Autonomous Racing Car on track at Silverstone during Formula Student UK Artificial Intelligence.
}
\vspace{-18pt}
  \label{fig:imeche}
\end{figure}

Autonomous racing competitions typically require a perception system to identify track boundaries, landmarks and other features. In this study, the vehicle (Fig. \ref{fig:imeche}) competes in the Institution of Mechanical Engineers (IMechE) Formula Student UK Artificial Intelligence competition\footnote{\url{https://www.imeche.org/events/formula-student/team-information/fs-ai}}\cite{case1996formula}, where 
events are characterized by colored traffic cones that demarcate different track layouts. These cones must be identified and mapped, enabling a path to be planned so that the vehicle can complete a lap at the highest possible safe speed.



\subsection{Identifying The Road Layout Ahead}  \label{sec:road-ahead}
For an autonomous race car, the ability to anticipate the upcoming road layout is fundamental for safe and optimal navigation. Unlike road vehicles operating at urban speeds, racing vehicles operate under conditions of high velocity, aggressive steering dynamics, and low tolerance for delay in decision-making. At racing speeds, small errors in perception can lead to large deviations in trajectory, resulting in loss of lap time or even safety-critical events.

To address this, we built a dedicated \textbf{TrackDrive Direction dataset} for this task (Sec. \ref{sec:trackdrive}). By identifying whether the road ahead is \textbf{\textit{straight}}, \textbf{\textit{left-hard}}, \textbf{\textit{left-medium}}, \textbf{\textit{left-easy}},\textbf{\textit{right-hard}}, \textbf{\textit{right-medium}}, or \textbf{\textit{right-easy}}, the vehicle gains the ability to:

\begin{itemize}
    \item Plan steering inputs in advance for unexpected turns.
    \item Optimize speed and acceleration, e.g., slowing earlier for sharper turns (left-hard/right-hard) while maintaining higher speeds on straights or gentler curves.
    \item Reduce uncertainty in navigation, since classification of the upcoming track provides the control module with discrete, interpretable categories.
    \item Downstream reasoning with uncertainty measures.
\end{itemize}

\textbf{Classifying the Road Layout} Our goal is to provide a high-level, discrete characterization of the upcoming track segment with uncertainty. Classification simplifies the interface between perception and control by mapping directly onto actionable decisions. For example, predicting `left-hard' signals the planner to prepare for aggressive steering and braking, while `straight' allows maximum throttle. 
Discrete predictions (e.g., `\textit{Left-Medium}' or sets like `\textit{\{Left-Medium, Left-Hard\}}') combined with uncertainty metrics (such as entropy) enable probabilistic reasoning and decision making. 

\textbf{Uncertainty management} is important as road layout prediction is inherently affected by occlusions, variable lighting, out of sample road layouts etc. By incorporating categorical predictions with uncertainty estimates, the vehicle can reason probabilistically, for e.g., slowing slightly if entropy is high between two similar classes. Also \textit{safety} is enhanced in autonomous racing, where dense cone layouts, narrow lanes, and changing conditions make accurate anticipation essential for avoiding surprises such as blind corners, thereby reinforcing robustness and overall performance.

\subsection{Using Uncertainty-aware Classifiers in the Autonomous Racing Car's Software Stack}  \label{sec:uq-av}

The AV control system pipeline comprises perception, path planning, localisation \& mapping and controls, implemented as individual nodes connected via ROS\footnote{\url{https://www.ros.org/}} \cite{macenski2022robot}. These components work in tandem to find the right actuation command to send to the VCU given the layout of the track (Fig. \ref{fig:ROS}).
\textit{\textbf{Perception}} is the entry point, as the car receives sensory information from the Zed 2i Stereo Camera and the Livox HAP TX LiDAR to identify cones in front of the vehicle. The data from the camera passes through the YOLOv11 Nano neural network \cite{wong2019yolo} to identify the class of each cone (blue, yellow, small orange, large orange). Data from these two sensors is fused to create an accurate map in cartesian space of the track. The uncertainty-aware classifier runs alongside this process, predicting the curvature of the track from data received from the camera. Data is collected from testing runs with the Autonomous racing car. The training data was captured at an interval of 5 frames per second from previous recordings from the sensors of the AV.
The positions of the cones are then passed to the localisation \& mapping and path planning submodules, which plan a trajectory for the vehicle to follow given the landmarks. This sends an actuation command to the system controller (i.e. speed request and a desired steering angle).

Before the system controller passes the actuation command to the VCU, it tunes the received speed request using the uncertainty of the upcoming turn received from the classifier\footnote{ROS-based code implementation will be released upon paper acceptance under an academic license.}, e.g., if a hard left turn is coming up but the classification of this turn from the classifier yields a high uncertainty, the speed request would be scaled down for the vehicle to excise caution until the classifier is sure that this is indeed a hard left turn, from which the correct speed request can be decided to navigate it.

\subsection{The TrackDrive Direction dataset} \label{sec:trackdrive}

To achieve reliable classification of road layouts, a dataset consiting of seven classes was created, comprising a direction and severity of the upcoming track:

\begin{itemize}
    \item \textbf{{Straight}}
    \item \textbf{{Left Turn}}: Left-easy, Left-medium, Left-hard
    \item \textbf{{Right Turn}}: Right-easy, Right-medium, Right-hard
\end{itemize}

This taxonomy reflects the natural variability in track curvature severity. Instead of attempting to regress continuous curvature values (which are sensitive to small annotation errors), this discrete categorization makes the task more tractable and robust.
In addition to the seven main classes, the dataset also includes three categories of \textbf{uncertain images} which are used for testing only: (1) \textbf{\textit{Random cone placement}}: cones placed randomly without forming a proper track direction, (2) \textbf{\textit{Fallen cones}}: cones lying on the track surface, creating ambiguity in the lane boundaries, and (3) \textbf{\textit{Confusing turns}}: layouts where cones form intersections or ambiguous curves, making the intended driving direction unclear.
These uncertain images are not used during training - instead they are solely for testing how well the model expresses uncertainty through measures such as entropy and mass functions assigned to sets in the presence of uncertainty.
In total, the dataset contains 1,187 labelled images, of which 919 are standard class samples (split into training/validation/test with 20\% of training data for validation) and 268 are uncertain cases reserved for testing model robustness. The dataset was collected by the racing team during multiple testing campaigns at two key locations: \textbf{Silverstone Circuit, UK}: https://www.silverstone.co.uk. \textbf{University Test Facilities}: Complementary data collected from controlled test tracks to balance samples and prevented model overfitting.

\begin{figure}[thpb]
    \centering
    \begin{subfigure}[b]{0.18\textwidth}
        \centering
        \includegraphics[width=\textwidth]{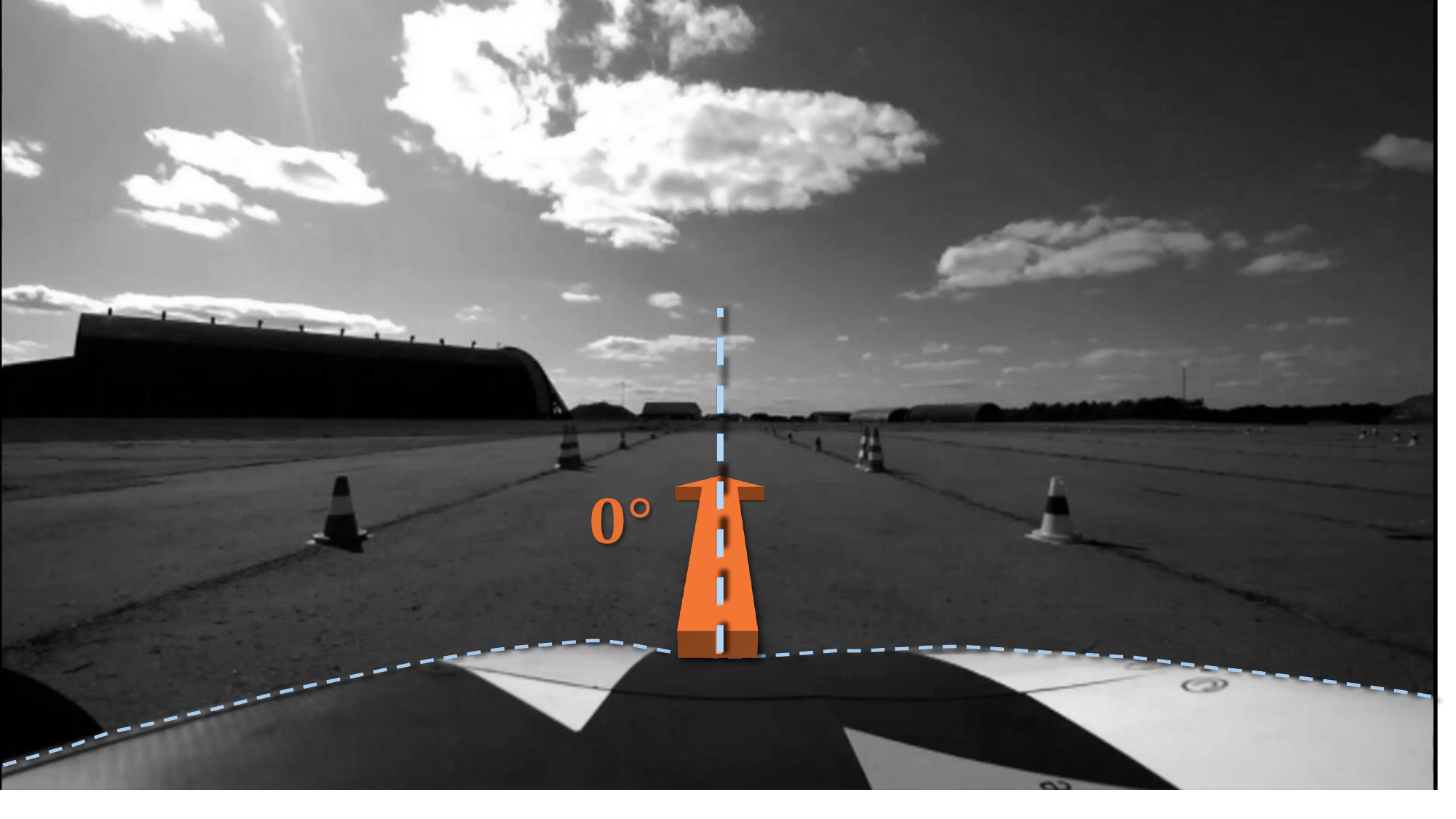}
        \caption{\textit{Straight}}
        \label{fig:straight}
    \end{subfigure}
    \begin{subfigure}[b]{0.18\textwidth}
        \centering
        \includegraphics[width=\textwidth]{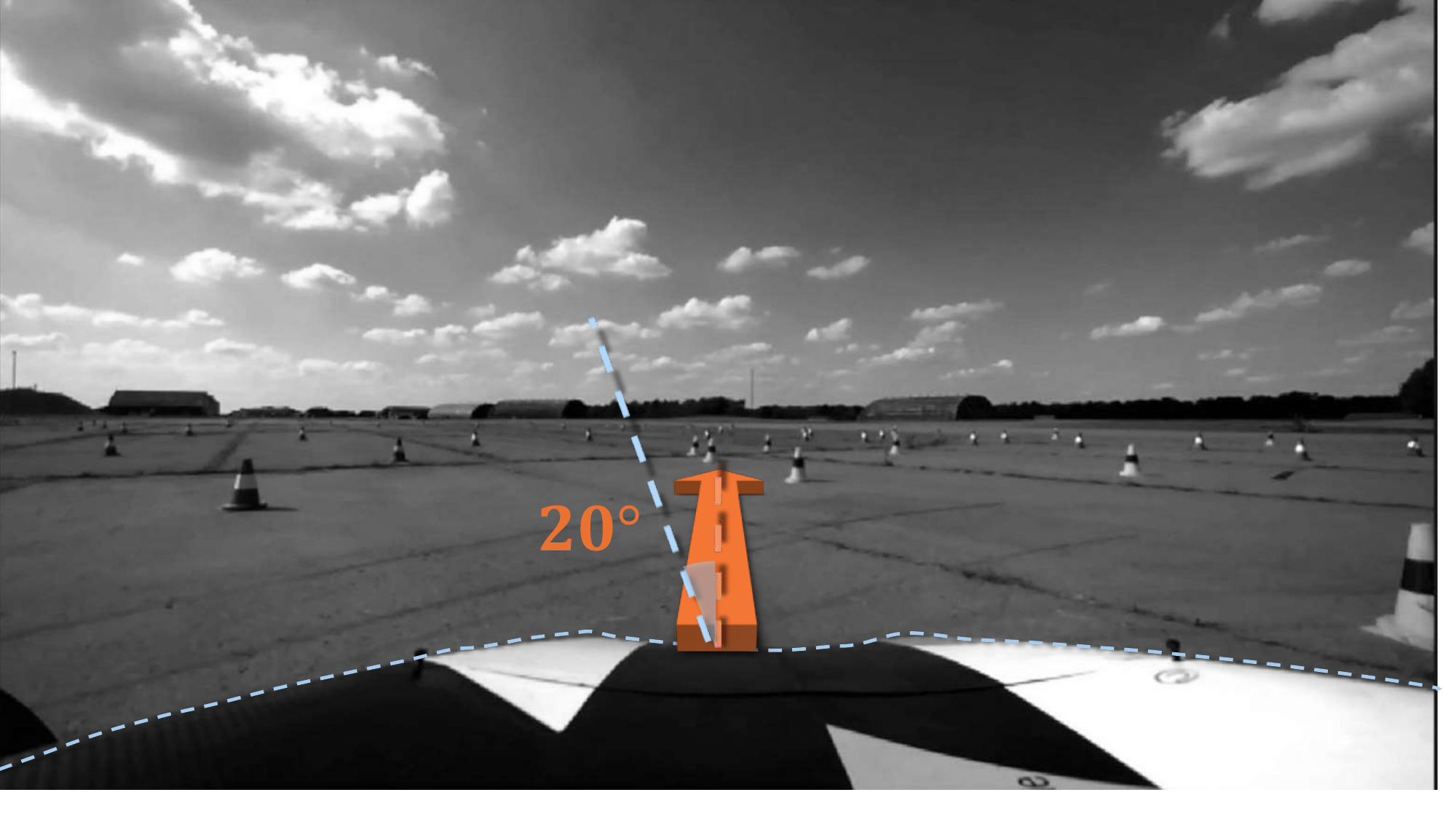}
        \caption{\textit{Left-Easy}}
        \label{fig:left_easy}
    \end{subfigure}
    

    \begin{subfigure}[b]{0.18\textwidth}
        \centering
        \includegraphics[width=\textwidth]{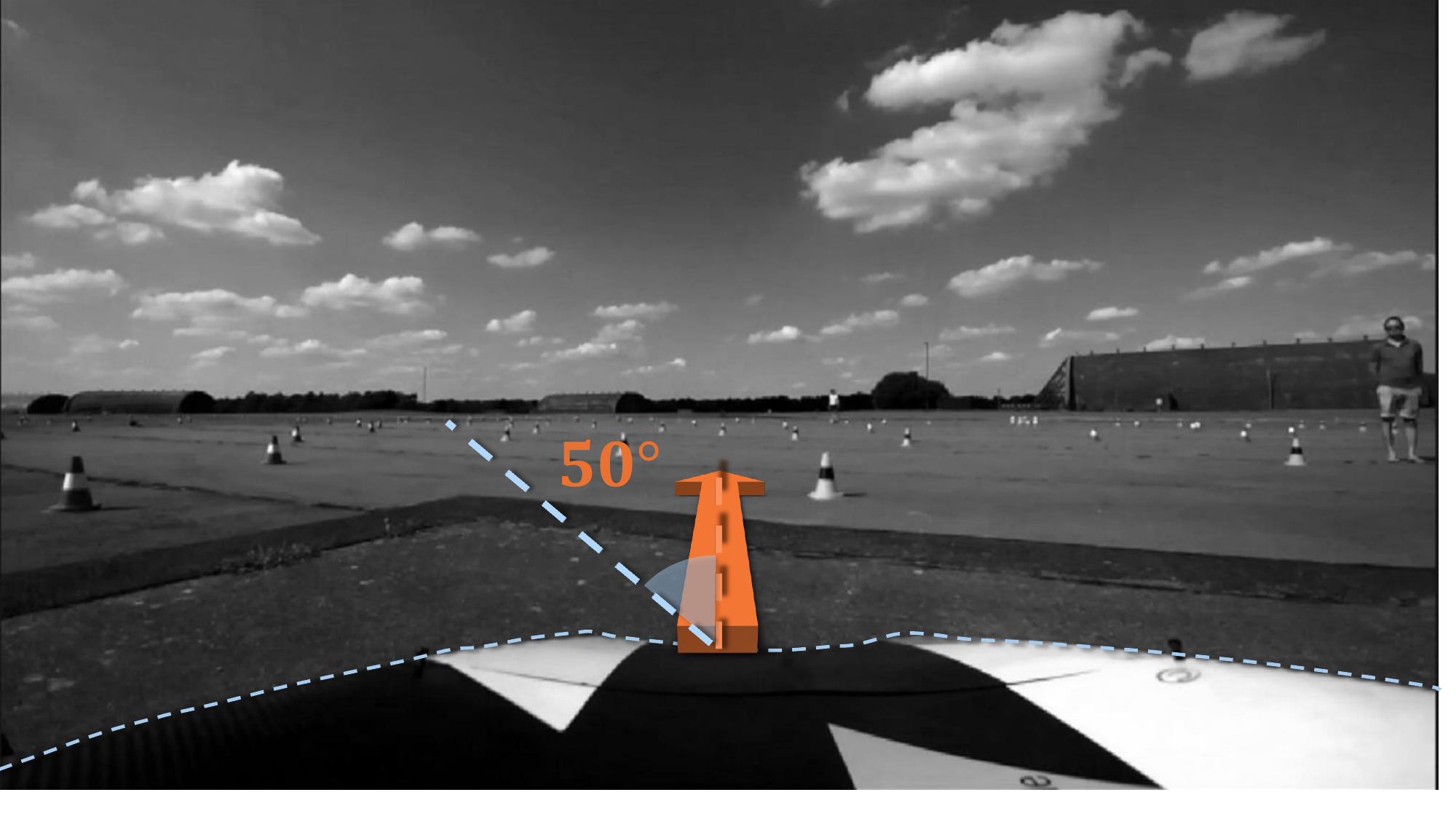}
        \caption{\textit{Left-Medium}}
        \label{fig:left_medium}
    \end{subfigure}
    \begin{subfigure}[b]{0.18\textwidth}
        \centering
        \includegraphics[width=\textwidth]{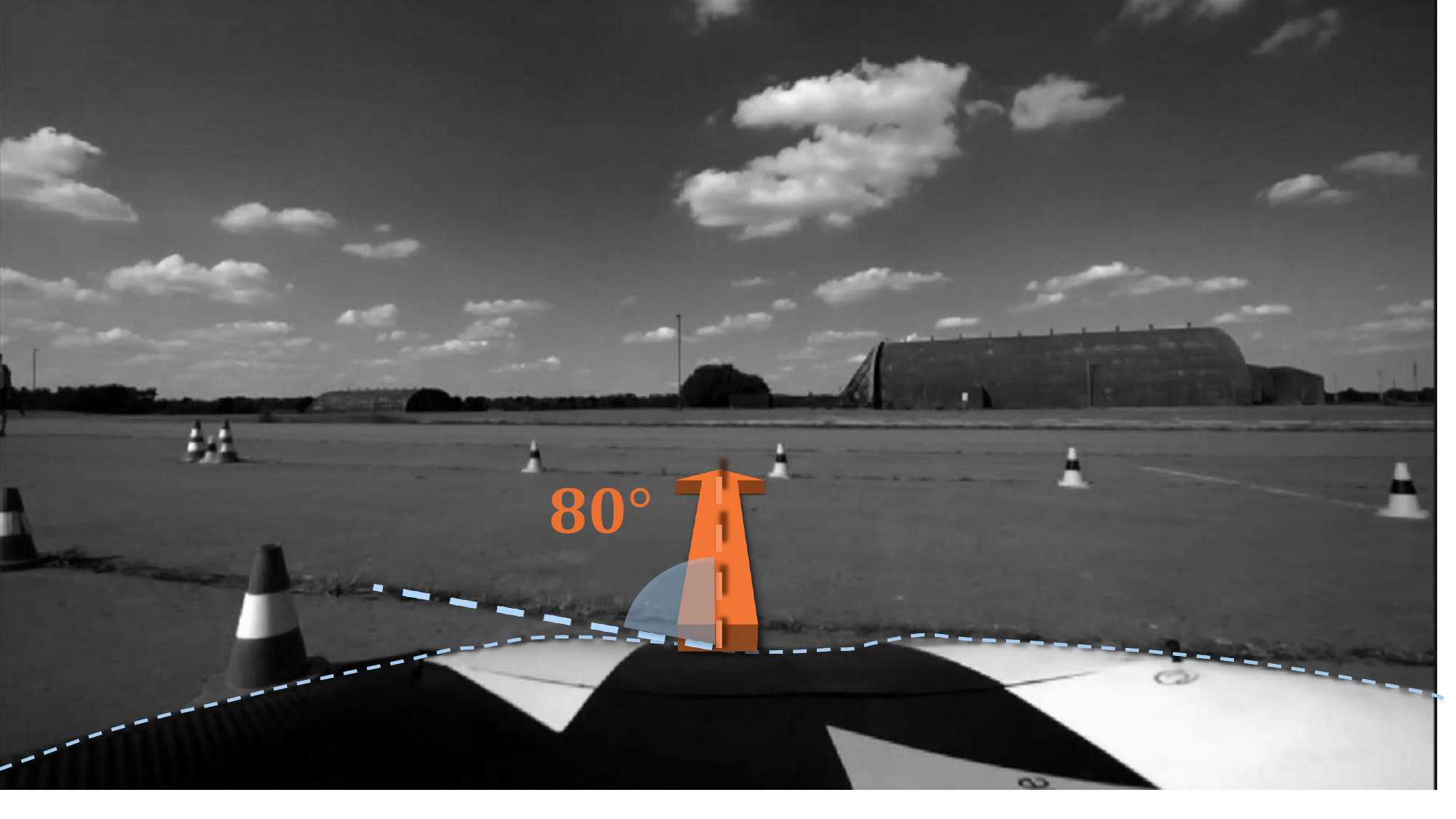}
        \caption{\textit{Left-Hard}}
        \label{fig:left_hard}
    \end{subfigure}
    \vspace{-4pt}
    \caption{Illustration of curvature classes: (a) \textit{Straight}, (b) \textit{Left-Easy}, (c) \textit{Left-Medium}, (d) \textit{Left-Hard}; right turns follow the same scheme. The orange arrow at the car's nose shows the forward axis, and the blue dashed line shows track deviation. Classes are defined by deviation: \textit{Straight} $<15^\circ$, Easy $15^\circ$–$35^\circ$, Medium $35^\circ$–$60^\circ$, Hard $>60^\circ$. Images are black-and-white to highlight geometry.}
    \vspace{-11pt}
    \label{fig:trackdrive-data}
\end{figure}

\textbf{Defining the Curvature Classes:}
To categorize turns as `easy', `medium', or `hard', we measure the track's deviation angle relative to the vehicle's forward direction (0°), represented by an orange arrow aligned with the car's nose (Fig. \ref{fig:trackdrive-data}). The boundaries are:

\begin{itemize}
    \item \textit{Straight}: deviation angle $<15^{\circ}$ (Fig. \ref{fig:straight}).
    \item Easy (Left/Right): $15^{\circ}– 35^{\circ}$ deviation (Fig. \ref{fig:left_easy}). Requires only mild steering adjustments.
    \item Medium (Left/Right): $35^{\circ}– 60^{\circ}$ deviation (Fig. \ref{fig:left_medium}). Noticeable curve requiring sustained steering effort.
    \item Hard (Left/Right): $>60^{\circ}$ deviation (Fig. \ref{fig:left_hard}). Represents sharp bends or near-90° turns requiring aggressive steering and deceleration.
\end{itemize}
  
\section{EXPERIMENTAL RESULTS} \label{sec:experiments}

We use a baseline \textit{CNN}, a \textit{RS-NN} \cite{manchingal2025random}, and a \textit{Laplace Bridge Bayesian Neural Network (LB-BNN)} \cite{hobbhahn2022fast}. The Trackdrive dataset comprises 919 labeled images of steering directions, with 735 for training (147 of these as validation) and 184 for testing. Additionally, 268 images labeled as `uncertain' contain ambiguous or challenging scenarios to assess model robustness. All images are 224×224 pixels.
\subsection{Comparative Analysis of Model Performance}   \label{sec:performance}
\textbf{Test accuracy:}
We evaluated CNN, RS-NN, and LB-BNN on the TrackDrive dataset with seven classes: \textit{Straight}, \textit{Left-Easy}, \textit{Left-Medium}, \textit{Left-Hard}, \textit{Right-Easy}, \textit{Right-Medium}, and \textit{Right-Hard}. Results (Tab.~\ref{tab:track-acc}) show clear performance differences: CNN achieved 67.93\%, RS-NN outperformed with 75.54\%, and LB-BNN lagged at 56.52\%. RS-NN’s gain highlights the benefit of explicit uncertainty modeling, while LB-BNN’s limited accuracy suggests overly conservative predictions on small data.

Ambiguous classes such as \textit{Right-Easy}, \textit{Right-Medium}, and \textit{Right-Hard} expose model weaknesses. CNN, relying on deterministic predictions, often misclassifies borderline samples. LB-BNN incorporates Bayesian parameter uncertainty but tends to under-generalize. RS-NN, in contrast, assigns belief masses across related sets (e.g., \textit{Right-Easy} with \textit{Right-Medium}, or \textit{Left-Medium} with \textit{Left-Hard}), reflecting structural similarities in data. This set-based treatment of uncertainty improves robustness, avoids overconfidence, excessive caution, and explains RS-NN’s superior accuracy. 

We examine how models perceive each class during training by analyzing active learning behavior (\S\ref{sec:AL}) under three acquisition strategies (\S\ref{sec:AL_1} to \S\ref{sec:AL_3}) revealing overall accuracy growth (Figs.~\ref{fig:AL-1},\ref{fig:AL-2},\ref{fig:AL-3}), per-class accuracy (Figs.~\ref{fig:AL-1-acc},\ref{fig:AL-2-acc},\ref{fig:AL-3-acc}), and per-class uncertainty (Figs.~\ref{fig:AL-1-unc},\ref{fig:AL-2-unc},\ref{fig:AL-3-unc}).

We tested a simplified 3-class version of the dataset: merging \textit{Left-Easy}, \textit{Left-Medium}, \textit{Left-Hard} into a single \textit{Left} class, and \textit{Right-Easy}, \textit{Right-Medium}, \textit{Right-Hard} into a single \textit{Right} class, leaving \textit{Straight}, \textit{Left}, and \textit{Right}. Accuracies (Tab.~\ref{tab:track-acc}) show RS-NN again leading (87.50\%), followed by LB-BNN (82.60\%) and CNN (81.06\%). The reduced performance gap reflects diminished class overlap after merging, and the gains for CNN and LB-BNN confirm that earlier struggles arose mainly from fine-grained distinctions in the seven-class task. All subsequent analyses are on the original 7-class scenario.

\vspace{-5pt}
\begin{table}[!htbp]
\caption{Test accuracy of CNN, RS-NN, and LB-BNN on the TrackDrive direction dataset. The \textbf{seven-class} scenario includes the classes: \textit{straight}, \textit{left-easy}, \textit{left-medium}, \textit{left-hard}, \textit{right-easy}, \textit{right-medium}, and \textit{right-hard}. The \textbf{three-class} scenario simplifies the classification task by merging directional classes into \textit{straight}, \textit{left}, and \textit{right}.}
\label{tab:track-acc}
\centering
\resizebox{0.9\linewidth}{!}{
\begin{tabular}{|c|c|c|}
\hline
\textbf{Model} & \textbf{Accuracy (7-class)} & \textbf{Accuracy (3-class)}\\
\hline
CNN    & $67.93 \pm 0.25$ & $81.06 \pm 0.05$ \\
\textbf{RS-NN}  & $\textbf{75.54} \pm \textbf{0.19}$ & $\textbf{87.50} \pm \textbf{0.19}$ \\
LB-BNN & $56.52 \pm 0.17$ & $82.60 \pm 0.23$ \\
\hline
\end{tabular}  }
\end{table}

\vspace{-5pt}
\subsection{Uncertainty Estimation}  \label{sec:uncertainty}

\begin{figure}[thpb]
    \centering
    \vspace{-6pt}
\includegraphics[width=0.45\textwidth]{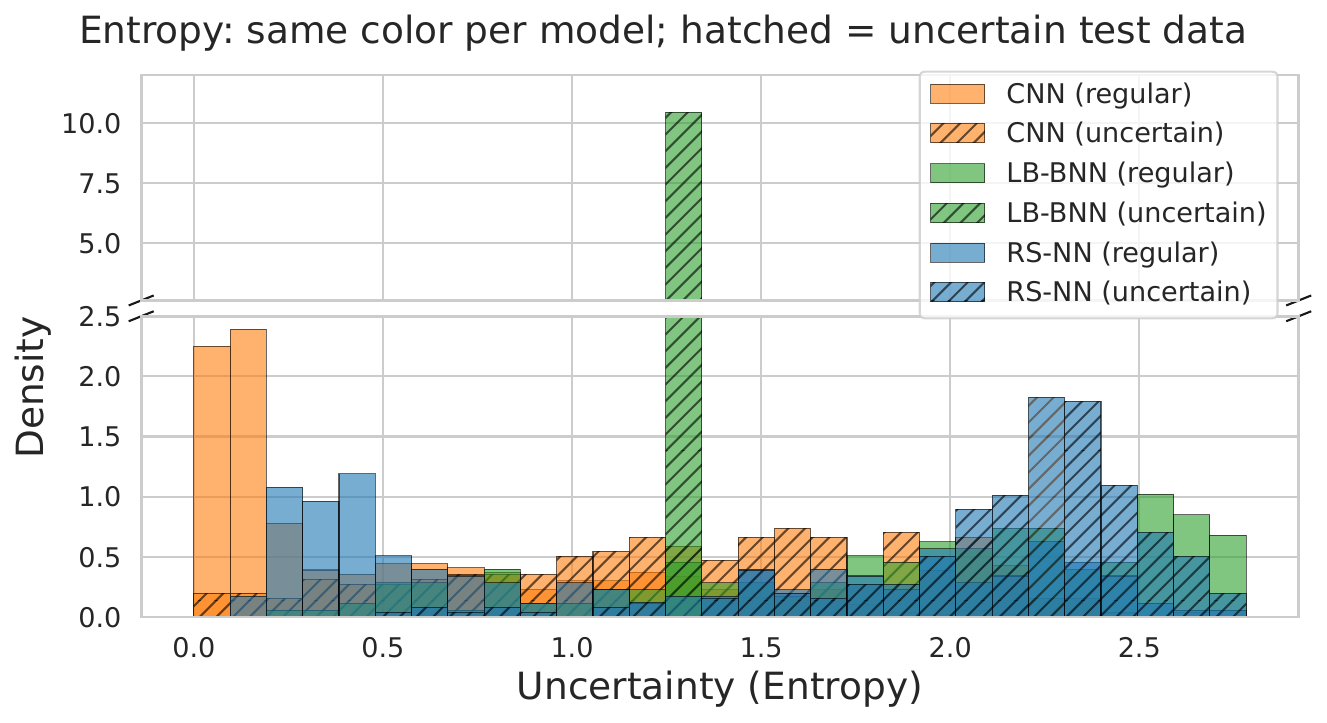} 
\vspace{-3pt}
    \caption{Entropy distributions: \textit{RS-NN}, \textit{CNN} and \textit{LB-BNN} on regular and uncertain (hatched) test data of Trackdrive dataset.
    }
    \vspace{-10pt}
    \label{fig:all-ent}
\end{figure}

Fig. \ref{fig:all-ent} shows a comparative perspective of uncertainty estimation across the three different models. Each model's predictions for regular test data are shown in solid colours and hatched bars represent predictions on uncertain test data. Notably, LB-BNN shows a pronounced spike at an entropy value near 1.3, suggesting a tendency towards conservative uncertainty estimates with moderate uncertainty for regular test samples. CNN entropy values are lower on average, implying generally confident but possibly overconfident predictions, reflected by lower entropy scores overall. Importantly, RS-NN presents a broader distribution of entropy scores across both regular and uncertain data. RS-NN entropy for uncertain data distinctly shifts toward higher values, clearly separated from regular data distributions, underscoring its effectiveness in identifying genuinely ambiguous scenarios. This ability to distinctly separate regular from uncertain predictions makes RS-NN particularly valuable for practical deployment where uncertainty-aware decision-making is crucial. RS-NN is the best performing model for accuracy and uncertainty, hence, we study the nature of its predictions in Sec. \ref{sec:confusion-rsnn} and apply it on a real-time AV in Sec. \ref{sec:real-results}.

\vspace{-0.04in}
\subsection{Confusion Analysis and Uncertainty Characterization in RS-NN Model Predictions}   \label{sec:confusion-rsnn}
\vspace{-0.04in}

We further analyze RS-NN performance by combining confusion matrices with entropy-based uncertainty estimation to identify strengths, weaknesses, and patterns of ambiguity. Key questions include:
\begin{itemize}
    \item Which classes are most frequently confused, and what does this reveal about model performance?
    \item Effectiveness of entropy capture prediction uncertainty?
    \item Is directional confusion linked to uncertainty?
\end{itemize}

\begin{figure}[thpb]
    \centering
    \vspace{5pt}
\includegraphics[width=0.45\textwidth]{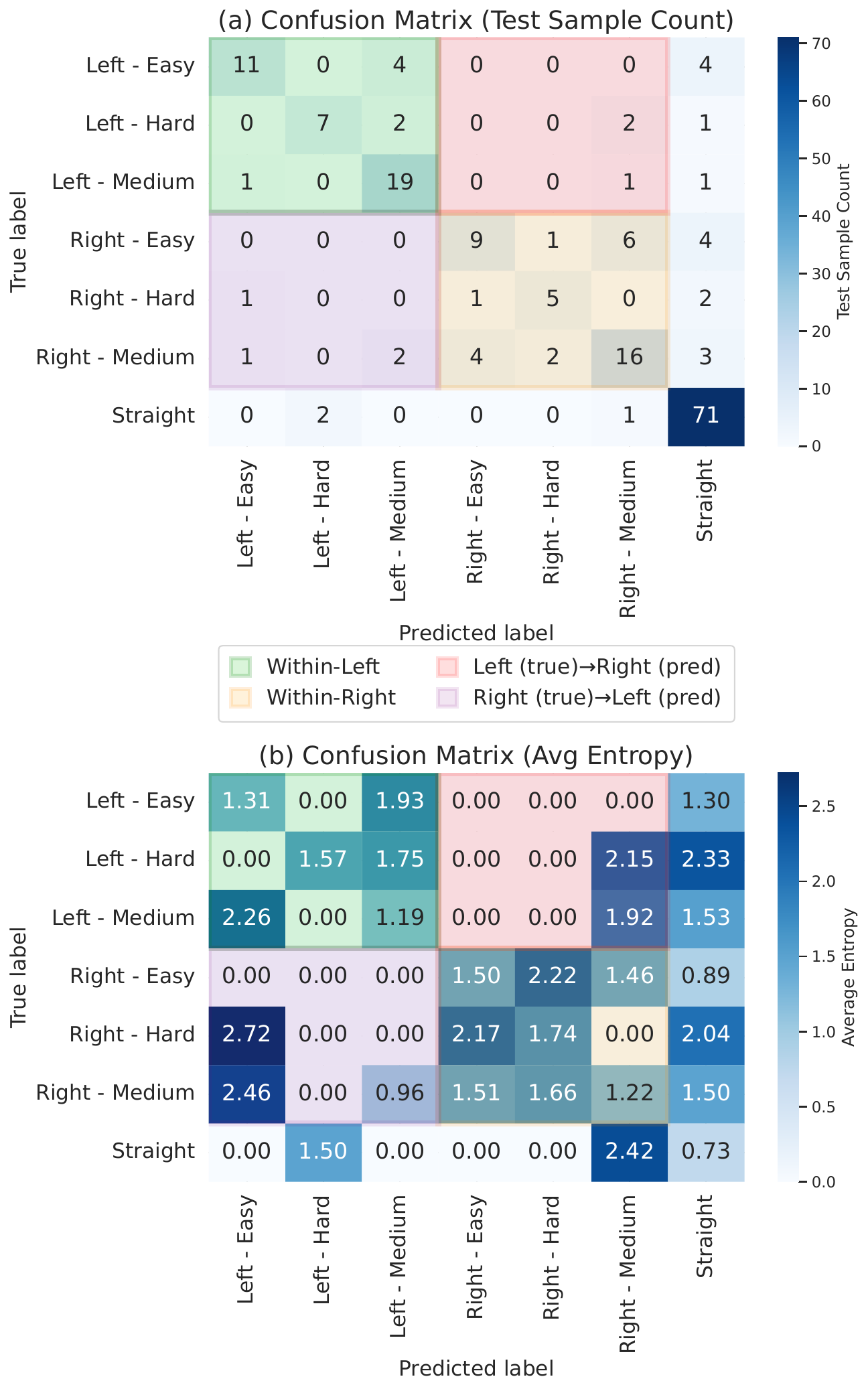} 
    \caption{Confusion matrices for RS-NN. (a) \textbf{Test sample counts:} frequency of correct and incorrect predictions across seven classes, with highlighted rectangles indicating within-group confusion (green, yellow) and between-group confusion (red, purple). (b) \textbf{Average entropy:} mean uncertainty for each true--predicted pair, where higher off-diagonal entropy reflects the inherent ambiguity and similarity between certain directional classes.}
\vspace{-10pt}
    \label{fig:conf-matrix}
\end{figure}

Fig.~\ref{fig:conf-matrix}a shows the confusion matrix in test sample counts. Diagonals indicate correct predictions, with \textit{Straight} most reliably recognized. Off-diagonals highlight in-directional confusions (\textit{Left-Easy} vs.~\textit{Left-Medium}, \textit{Right-Easy} vs.~\textit{Right-Medium}), marked by green/yellow rectangles. Less frequent inter-directional confusions (\textit{Left}$\rightarrow$\textit{Right}, \textit{Right}$\rightarrow$\textit{Left}), shown in red/purple rectangles, reflect natural ambiguity in edge cases. Fig.~\ref{fig:conf-matrix}b complements this with pignistic entropy for each true-predicted pair. Low entropy values align with confident, correct predictions (e.g., \textit{Straight} with mean entropy 0.61). Higher entropy accompanies ambiguous cases, such as \textit{Right-Hard}$\rightarrow$\textit{Left-Easy} (2.31) or \textit{Right-Hard}$\rightarrow$\textit{Right-Medium} (2.45). This confirms entropy as an effective quantitative measure of uncertainty.

Together, matrices reveal that RS-NN attain high accuracy and low uncertainty in distinct cases (\textit{Straight}), but struggles with intra-directional subtleties (e.g., \textit{Easy} vs.~\textit{Medium} turns). Entropy reliably highlights these ambiguous regions, showing uncertainty is more pronounced in directions than across them. These insights suggest that RS-NN’s main challenge lies in distinguishing fine-grained turn variations.

\begin{figure*}[thpb]
    \centering
    \hspace{-0.6em}
        \includegraphics[width=0.6\textwidth]{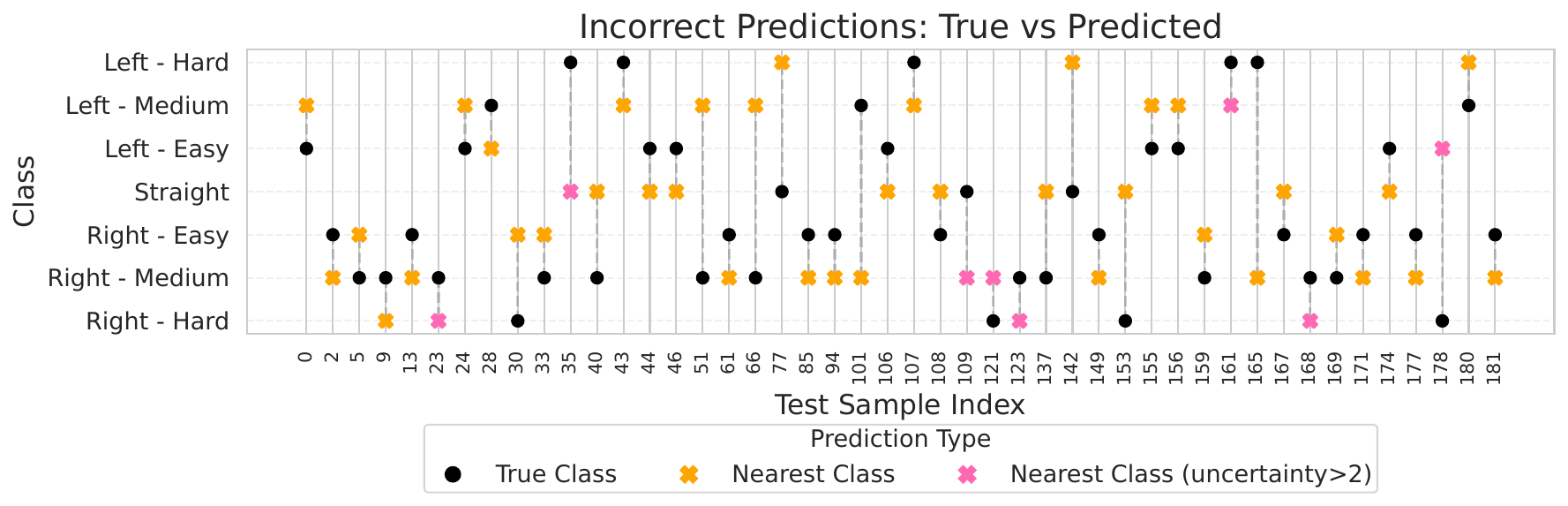} 
    \caption{Visualization of RS-NN misclassifications: black dots: true classes, crosses: nearest predictions (orange: low uncertainty, pink: high uncertainty). Dashed lines illustrate confusion between closely related classes.
    }
      \vspace{-15pt}
    \label{fig:incorrect-pred}
\end{figure*}

Fig.~\ref{fig:incorrect-pred} shows incorrect predictions, confirming most errors occur in closely related directional classes (e.g., \textit{Right-Medium}$\leftrightarrow$\textit{Right-Easy}, \textit{Left-Easy}$\leftrightarrow$\textit{Left-Medium}), further illustrating subtle feature similarities driving misclassification.

Black dots represents the true class of a test sample, crosses indicate the nearest predicted class in the top mass set. The nearest class is determined by identifying the closest class included in the top mass set predicted by the RS-NN model. Dashed grey lines illustrate the proximity and confusion between closely related classes. The distinction between low uncertainty (orange crosses) and high uncertainty predictions (pink crosses) highlights the model's capacity to recognise and quantify ambiguity effectively. High uncertainty cases (entropy $>$ 2) underscore the difficulty of making precise classifications in particularly ambiguous scenarios. 

\subsection{Using RS-NN in Autonomous Racing Vehicle Control: Results}   \label{sec:real-results}

To validate RS-NN under realistic driving conditions, we integrated our uncertainty-aware classifier into the vehicle's software pipeline via ROS. The classifier runs inside of a dedicated ROS node which subscribes to the forward-facing camera, runs the RS-NN inference pipeline in real time, and publishes both the predicted Trackdrive direction class and its entropy on the predicted direction. 

\begin{figure}[thpb]
    \centering
    \vspace{5pt}
\includegraphics[width=0.4\textwidth]{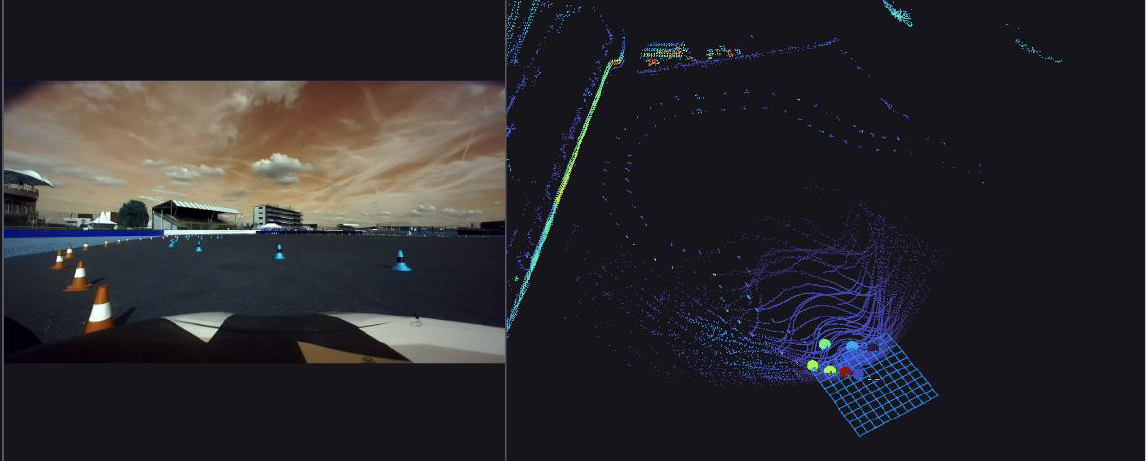} 
    \caption{Foxglove Test Dashboard: camera video (left), LIDAR scan (right).
    }
      \vspace{-15pt}
    \label{fig:foxglove-dash}
\end{figure}

Fig. \ref{fig:foxglove-dash} showcases our real‐time testing setup for the RS-NN model using Foxglove Studio (\url{https://www.foxglovestudio.co.uk/}). 
Left panel has live camera feed. Right panel displays a 3D LiDAR scan, with detected cones rendered as colored spheres.

The direction of the upcoming road layout is classified using the epistemic classifier - and the unique ability of RS-NN to `Know When it Does Not Know' is then leveraged to intervene in uncertain scenarios - modulating the speed request in response to areas of high uncertainty in the track layout.

A five‐tier speed‐scaling policy was implemented, based on the RS-NN model's predicted entropy at each timestep:
\begin{itemize}
    \item Entropy$<$2.2: run at full requested velocity (RPM\footnote{Wheel RPM: (revolutions per minute) is the rotational speed of a vehicle’s wheel, typically measured by wheel-speed sensors.}).
    \item 2.2$<$Entropy$<$2.3: reduce velocity to 90\% (slight slowdown).
    \item 2.3$<$Entropy$<$2.4: reduce to 80\% (moderate slowdown).
    \item 2.4$<$Entropy$<$2.6: reduce to 60\% (stronger slowdown).
    \item Entropy$\ge$2.6: cut throttle entirely (full stop).
\end{itemize} 
 
\begin{figure}[thpb]
    \centering
    \vspace{7pt}
\includegraphics[width=0.4\textwidth]{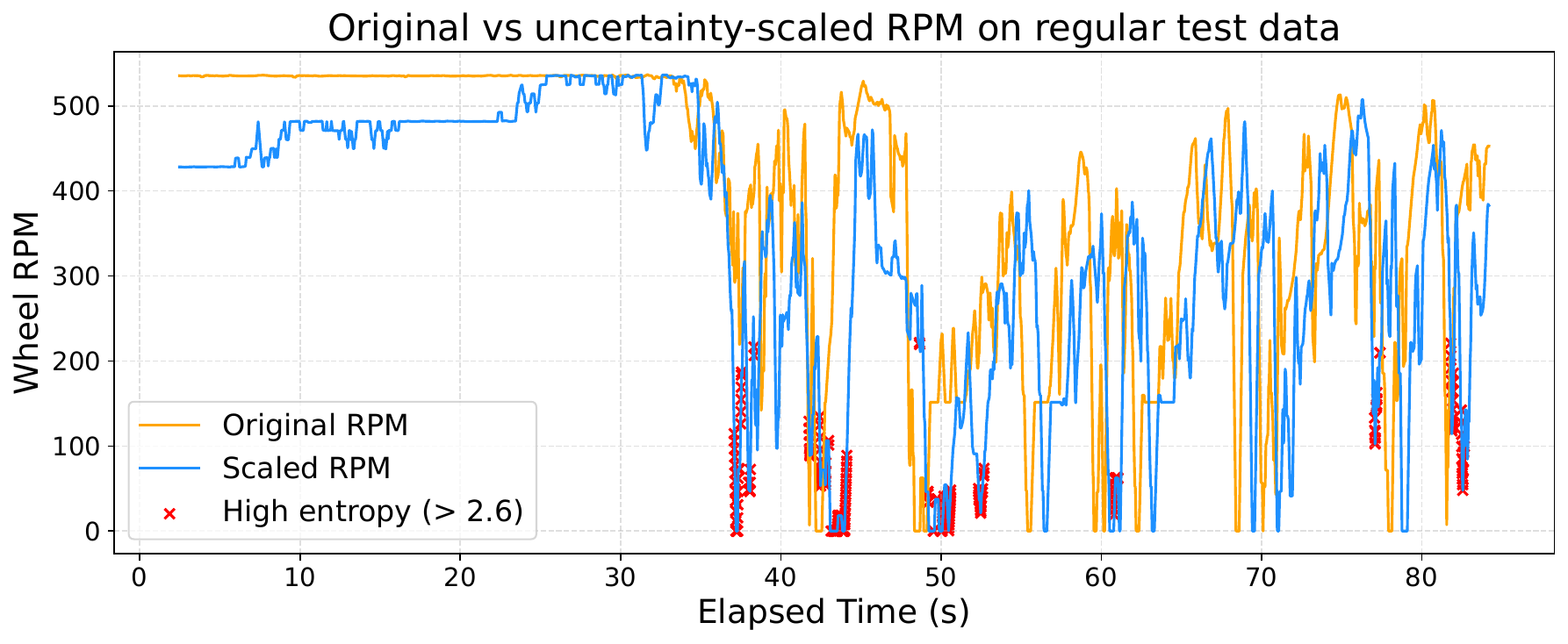} 
        \hspace*{1em}
        \includegraphics[width=0.4\textwidth]{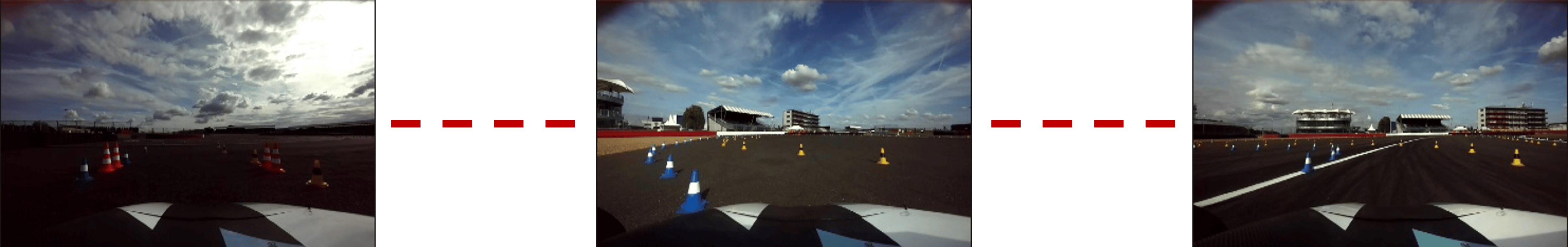} 
    \caption{ Velocity traces for original vs. uncertainty-scaled RPM on regular track drive. Orange is the planner's requested RPM; blue is the uncertainty‐scaled RPM. 
    }
      \vspace{-12pt}
    \label{fig:realtime-reg}
\end{figure} 

In Fig. \ref{fig:realtime-reg}, two throttle‐related topics are overlaid in real time during a lap. The orange trace is the original RPM command from the high‐level planner, while the blue trace is the adjusted RPM after applying our RS-NN uncertainty‐based throttle controller. High-entropy moments where the RS-NN classifier reports entropy $>$ 2.6 are highlighted with red `x' markers. The plot shows that on straightaways and gentle curves (the first $\approx$35s), entropy remains low, so the scaled RPM closely follows the original command. As the car approaches more complex turns ($\approx$35-50s), isolated red crosses indicate brief spikes in uncertainty; here the blue line dips below the orange, enforcing slight slowdowns. Later in the run ($\approx$50-80s), clusters of high-entropy points correspond to tighter or more ambiguous sections, triggering stronger RPM reductions and even momentary near-stops. Overall, the plot shows that on a familiar, regular course, uncertainty-driven scaling introduces targeted slowdowns only when and exactly where the model is least confident, otherwise preserving full throttle. This dynamic interplay clearly demonstrates how real‐time uncertainty estimation can directly modulate vehicle behaviour, trading off speed for safety exactly when and where the perception model is least certain.

\begin{figure}[thpb]
    \centering
        \vspace{3pt}
\includegraphics[width=0.43\textwidth]{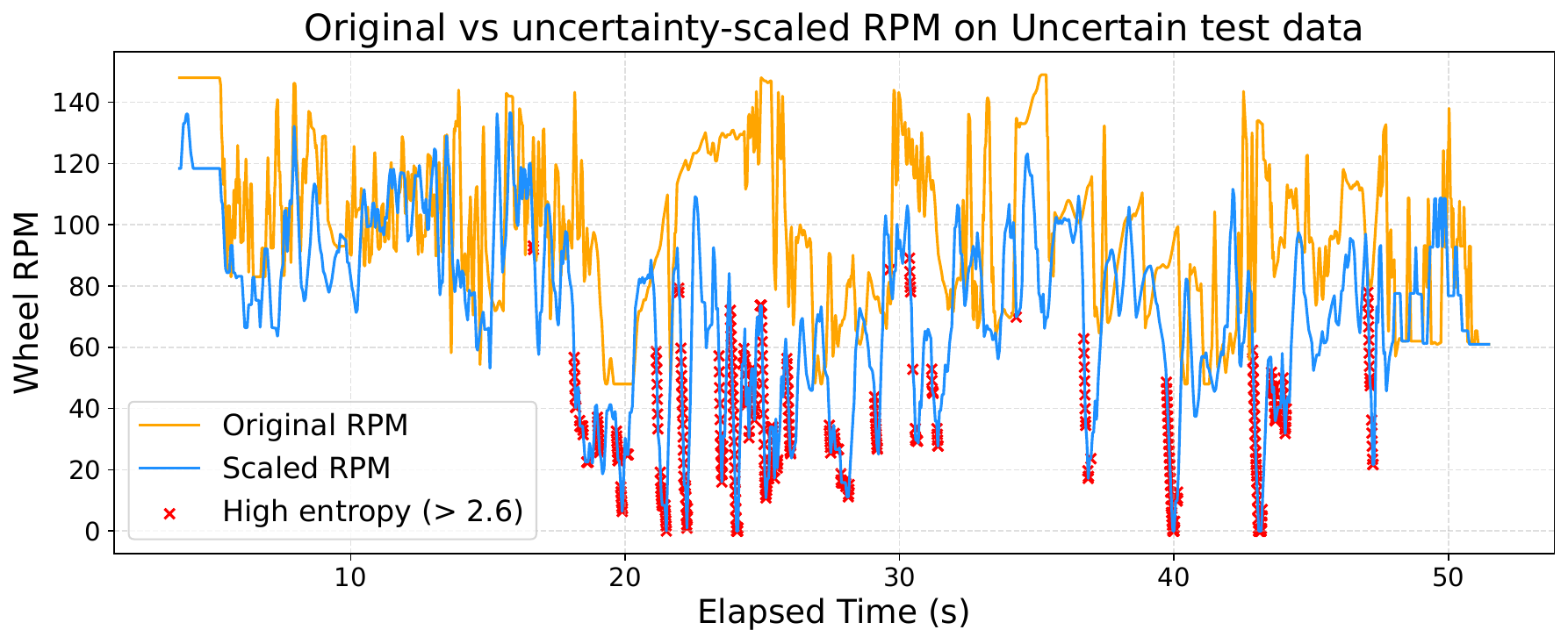} 
        \hspace*{1em}
        \includegraphics[width=0.43\textwidth]{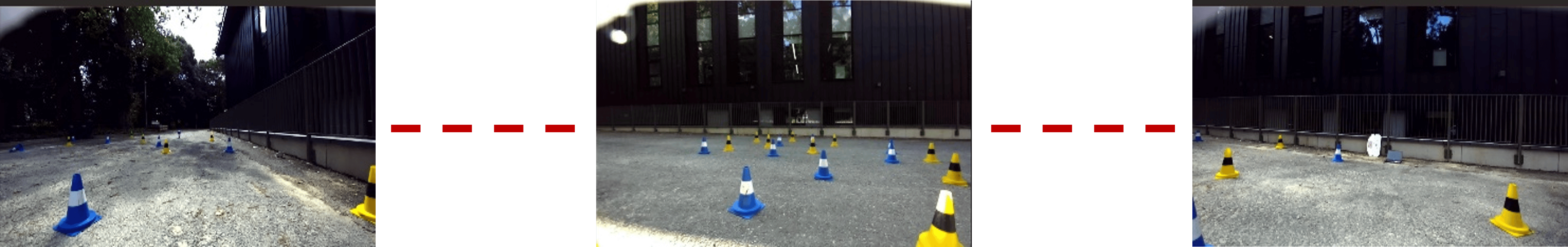} 
    \caption{Velocity traces for original vs. uncertainty-scaled RPM on uncertain (Random/Fallen Cone) track drive. Orange: planner's requested RPM; Blue: uncertainty‐scaled RPM.}
    \label{fig:realtime-unc}
        \vspace{-15pt}
\end{figure}  

Fig. \ref{fig:realtime-unc} shows the same RPM comparison, but on a deliberately uncertain test track featuring random cone placements and fallen cones. Here, the classifier's entropy frequently exceeds 2.6 (dense red `×' clusters between $\approx$15–45 s), signalling pervasive ambiguity in track geometry. Consequently, the uncertainty-scaled RPM (blue) is substantially and persistently lower than the original commands (orange), with frequent drops to near zero when entropy peaks. Unlike the smooth regular course, the scaled throttle now enforces extended slowdowns throughout the run, reflecting the system's cautious response to unfamiliar or misleading cone layouts.

By overlaying the blue and orange signals, we can clearly see the controller's safety envelope in action: proactive slowdowns or even stops in the face of uncertain perception, and full-speed operation when confidence recovers. 
This demonstrates how entropy-based control can translate model uncertainty into tangible, conservative driving behaviour; prioritizing safety whenever perceptual confidence is compromised.

\subsection{Active Learning}   \label{sec:AL}
	
Active Learning (AL) improves model performance while reducing labelling costs by strategically selecting the most informative samples, often based on prediction uncertainty, rather than random sampling. This is particularly effective for uncertainty estimation, as AL targets ambiguous or borderline cases, improving decision boundary learning and model robustness, critical in safety-sensitive domains \cite{10610405}. Our experiments show that the RS-NN achieves superior results with less data, and we evaluate its performance through three active learning experiments using different initial sampling strategies and acquisition methods.

\subsubsection{Experiment 1: Stratified Initial Sampling with Uncertainty-based Acquisition} \label{sec:AL_1}
\begin{figure}[!t]
    \centering
    \begin{minipage}[t]{0.49\textwidth}
        \centering
        \includegraphics[width=\textwidth]{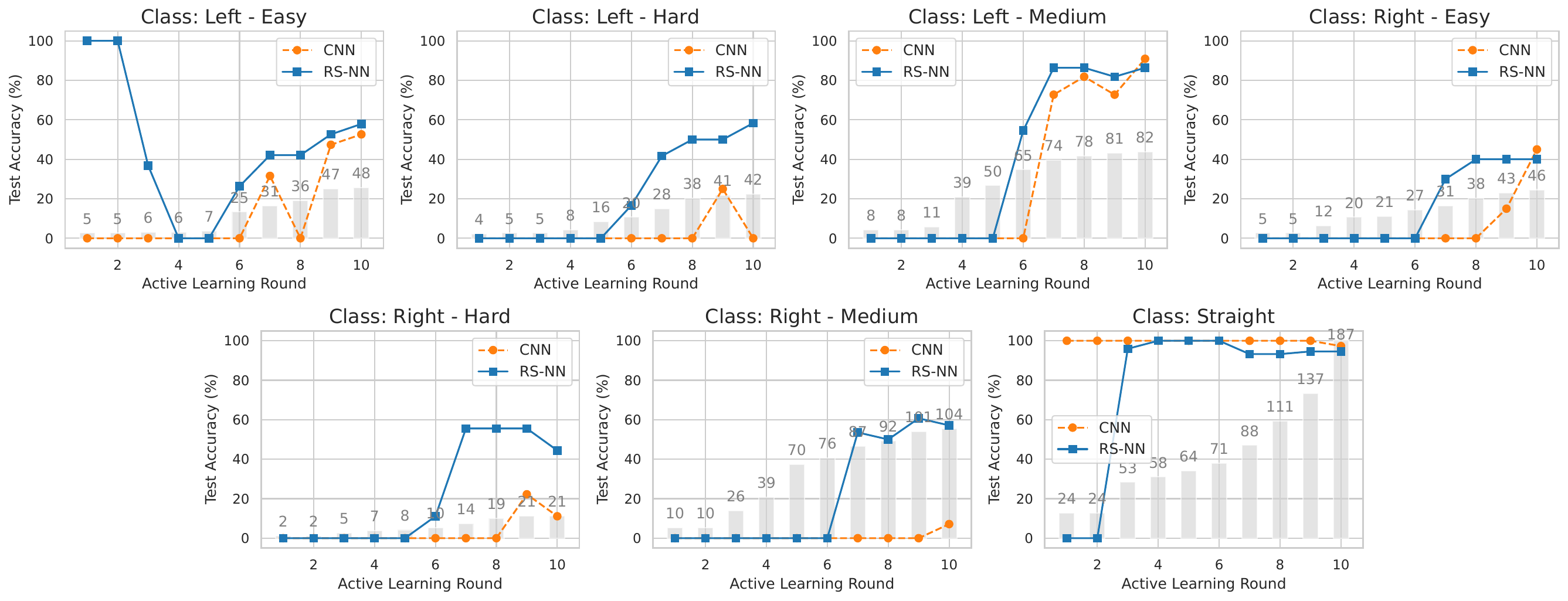}
        \caption{\textit{Experiment 1:} Per-class RS-NN vs CNN accuracy across active-learning rounds. \textit{Straight} remains high from the start, while minority classes (e.g., \textit{Left-Hard}, \textit{Right-Easy}) lag early due to imbalance but improve as uncertain samples are added.}
         \vspace{-13pt}
        \label{fig:AL-1-acc}
    \end{minipage}
    \end{figure}

We seed training with 10\% stratified samples, reflecting natural class imbalance (dominated by Straight). This leads to initial overfitting on majority classes and poor minority performance. Subsequent acquisitions use entropy-based uncertainty, adding the most informative samples without correcting class imbalance.
\begin{figure}[thpb]
    \centering
\vspace{3pt}        \includegraphics[width=0.28\textwidth]{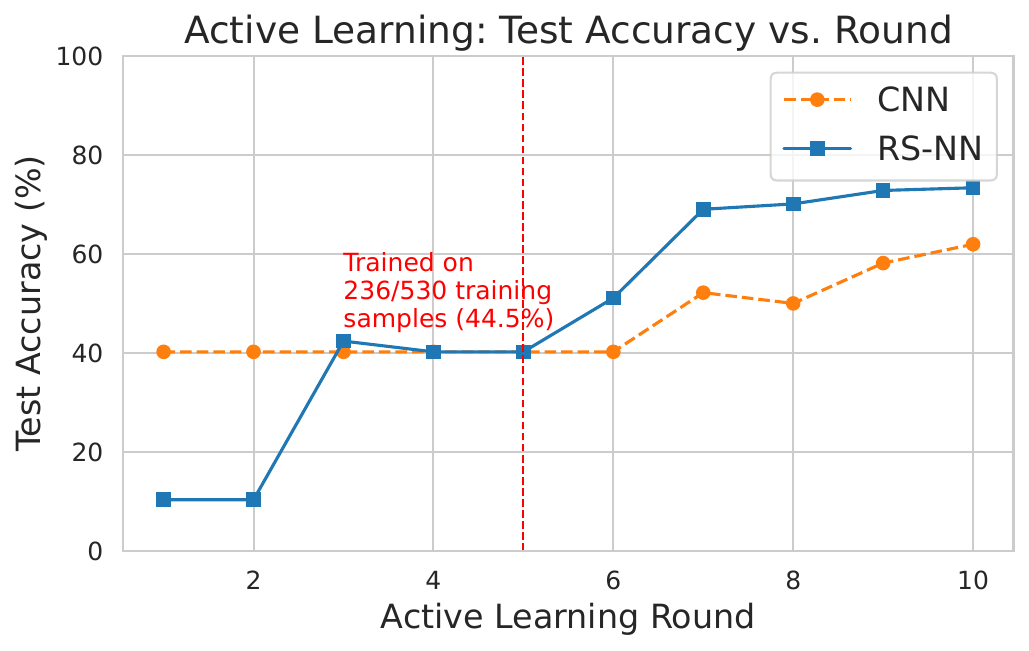} 
\vspace{-3pt} 
    \caption{\textit{Experiment 1:} Test accuracy with 10\% seeding: CNN improves slowly from ~40\%, while RS-NN climbs faster and surpasses it by leveraging uncertainty.}
      \vspace{-10pt}
    \label{fig:AL-1}
\end{figure}
Results (Fig. \ref{fig:AL-1}): Both models start with low accuracy, but RS-NN quickly outpaces CNN after ~45\% of data is labelled, highlighting the value of uncertainty-driven sampling. Class-level plots show Straight achieves high accuracy early, while minority classes improve later as more uncertain samples are added; RS-NN consistently outperforms CNN.
The group of plots in Fig. \ref{fig:AL-1-acc} presents the accuracy progression per class through the active learning rounds. Each subplot tracks one class: `\textit{Straight}' achieves near-perfect accuracy immediately, reflecting its dominance in the seed set and its distinct visual features. In contrast, minority turn classes (e.g., \textit{Left-Hard}, \textit{Right-Easy}) lag early on and improve only in later rounds as more uncertain, class-specific samples are acquired. In every case, RS-NN (blue) outperforms the CNN (orange).

\begin{figure}[!t]
    \begin{minipage}[t]{0.49\textwidth}
        \centering
        \includegraphics[width=\textwidth]{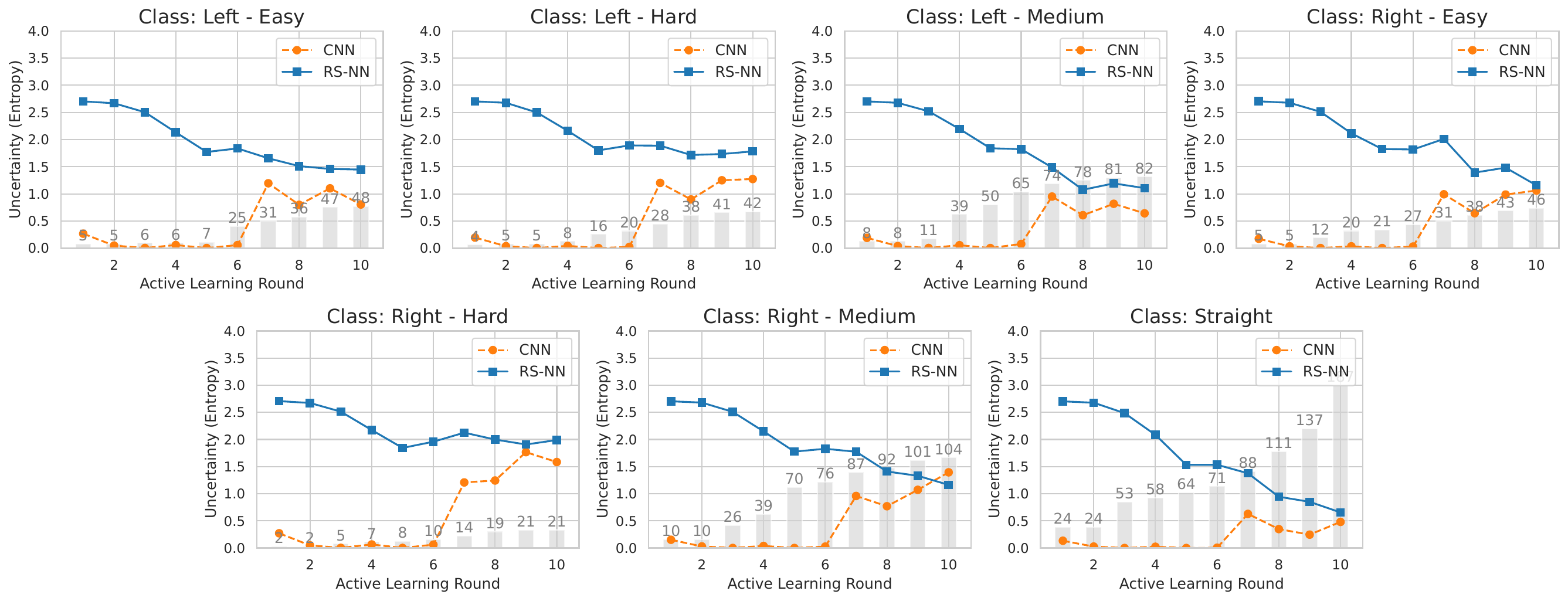}
        \caption{\textit{Experiment 1:} Per-class RS-NN vs CNN uncertainty (mean entropy) across active-learning rounds. Uncertainty declines fastest for majority classes like \textit{Straight}, while ambiguous turn classes (e.g., \textit{Left-Medium}, \textit{Right-Hard}) remain higher until more targeted samples are acquired.}
          \vspace{-18pt}
        \label{fig:AL-1-unc}
    \end{minipage}
\end{figure}

Uncertainty trends (Fig. \ref{fig:AL-1-unc}): RS-NN initially shows higher entropy, reflecting better-calibrated uncertainty, which decreases steadily as training progresses. CNN shows misplaced confidence early (low entropy despite poor accuracy), with entropy rising only later for harder classes. By the final round, RS-NN achieves both higher accuracy and lower uncertainty.





\subsubsection{Experiment 2: Balanced Initial Sampling with Class-wise Acquisition}   \label{sec:AL_2}

\begin{figure}[thpb]
    \centering
        \vspace{-5pt}\includegraphics[width=0.3\textwidth]{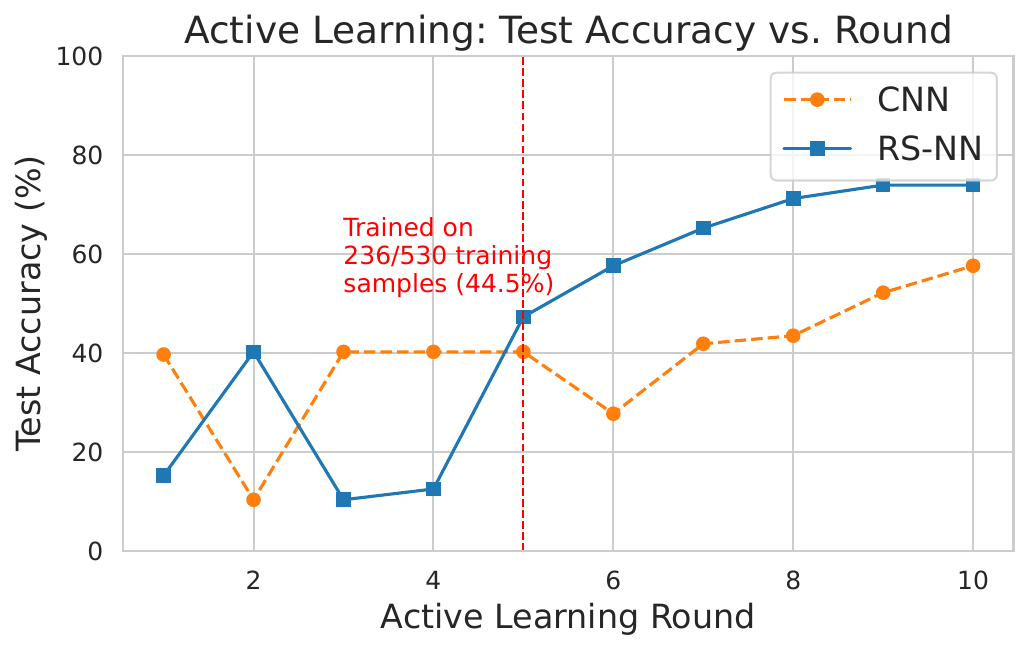} 
        \vspace{-3pt} 
    \caption{\textit{Experiment 2:} Test‐set accuracy RS-NN vs CNN as a function of the number of labelled examples under a balanced seed and class-wise acquisition.
    }
    \vspace{-18pt}
    \label{fig:AL-2}
\end{figure}

\begin{figure}[!h]
    \centering
        \centering
        \includegraphics[width=0.49\textwidth]{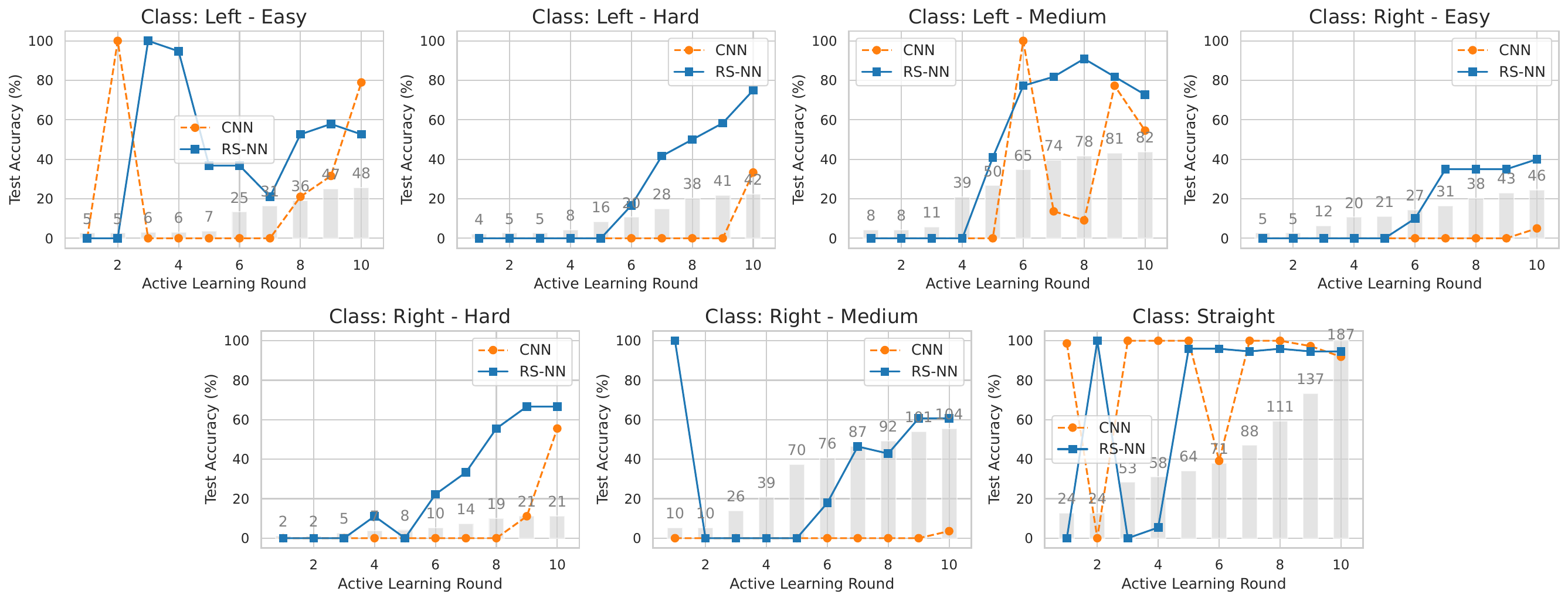}
        \caption{\textit{Experiment 2:} Per-class accuracy  of RS-NN vs CNN under balanced acquisition.}
          \vspace{-13pt}
        \label{fig:AL-2-acc}
    \end{figure}

\begin{figure}[!h]
\vspace{-15pt}
        \centering
        \includegraphics[width=0.49\textwidth]{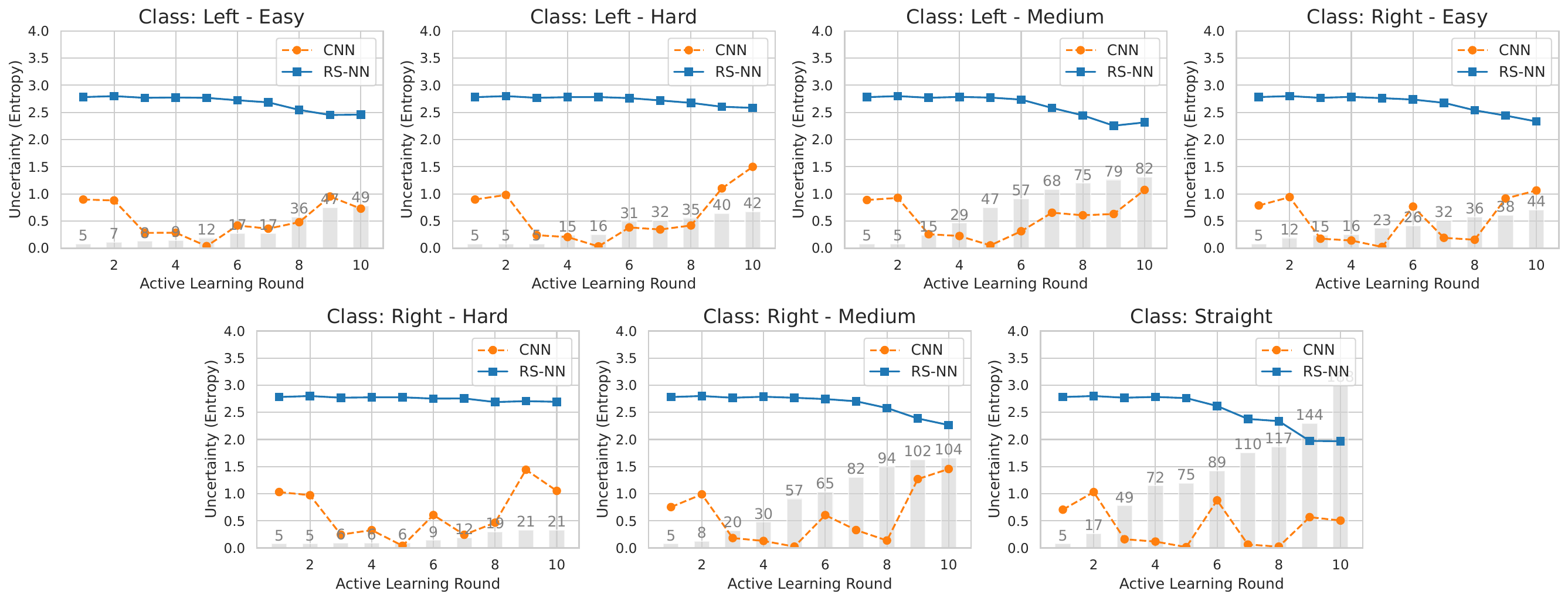}
        \caption{\textit{Experiment 2:} Per-class uncertainty (entropy) of RS-NN vs CNN under balanced acquisition.}
        \vspace{-13pt}
        \label{fig:AL-2-unc}
\end{figure}

This experiment addresses the imbalance of Experiment~\ref{sec:AL_1} by starting with a balanced seed (five samples per class) and continuing with per-class uncertainty-based acquisition, ensuring equal representation across all classes. This strategy reduces bias, improves robustness, and enhances accuracy and uncertainty estimation. Under this balanced regime, RS-NN rapidly leverages uncertainty to outperform CNN (Fig.~\ref{fig:AL-2}). Starting near 40\%, RS-NN’s accuracy rises to $\sim$70\% by 5 and plateaus near 73\%, while CNN improves slowly and never catches up. This highlights RS-NN’s ability to efficiently exploit informative per-class samples. Class-level accuracy (Fig.~\ref{fig:AL-2-acc}) shows minority classes like \textit{Left-Hard} and \textit{Right-Hard} improving steadily from near-zero to above 50\%, while majority classes (\textit{Straight}, \textit{Left-Medium}) achieve high accuracy quickly but still gain from targeted acquisitions. Grey bars confirm each class receives equal new labels per round. Uncertainty trends (Fig.~\ref{fig:AL-2-unc}) reveal RS-NN starts with higher entropy, particularly for ambiguous turns, reflecting better-calibrated caution. As per-class uncertain samples are added, entropy consistently declines alongside accuracy. CNN, however, shows misplaced early confidence (low entropy despite poor accuracy), only exhibiting higher entropy in later rounds. Overall, RS-NN demonstrates faster learning and superior calibration under active learning.

 

\begin{figure}[t]
    \centering
        \includegraphics[width=0.3\textwidth]{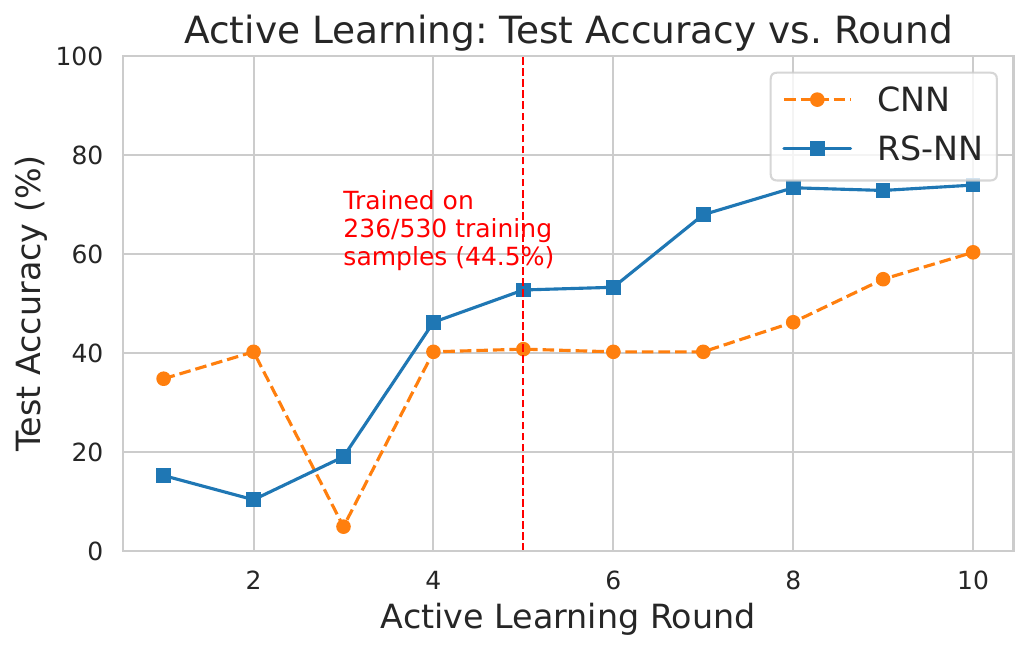} 
        \vspace{-3pt} \caption{\textit{Experiment 3:} Test accuracy vs.\ active-learning round under fixed cumulative class targets (5 labels per class added each round). RS-NN (blue) consistently surpasses CNN (orange) by leveraging uncertainty sampling in a fully balanced schedule.}
\vspace{-10pt}
    \label{fig:AL-3}
\end{figure}

\begin{figure}[!h]
    \centering
    \includegraphics[width=0.49\textwidth]{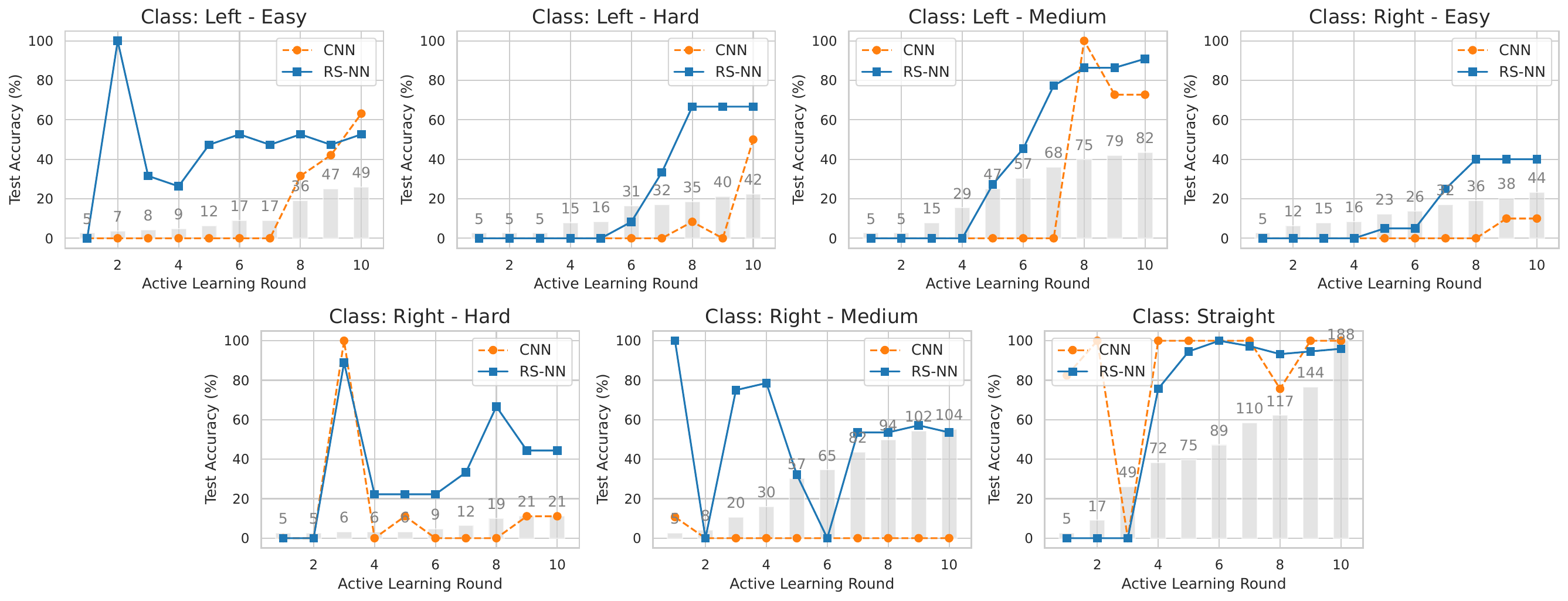}
        \caption{\textit{Experiment 3:} Per-class test accuracy under Experiment~3's fixed per-class labelling schedule. Grey bars show labelled samples per class; accuracy for all classes improves steadily and predictably as more data are added equally.}
        \vspace{-18pt}
        \label{fig:AL-3-acc}
    \end{figure}
    \begin{figure}[!h]
    \vspace{-15pt}
        \centering
    \includegraphics[width=0.49\textwidth]{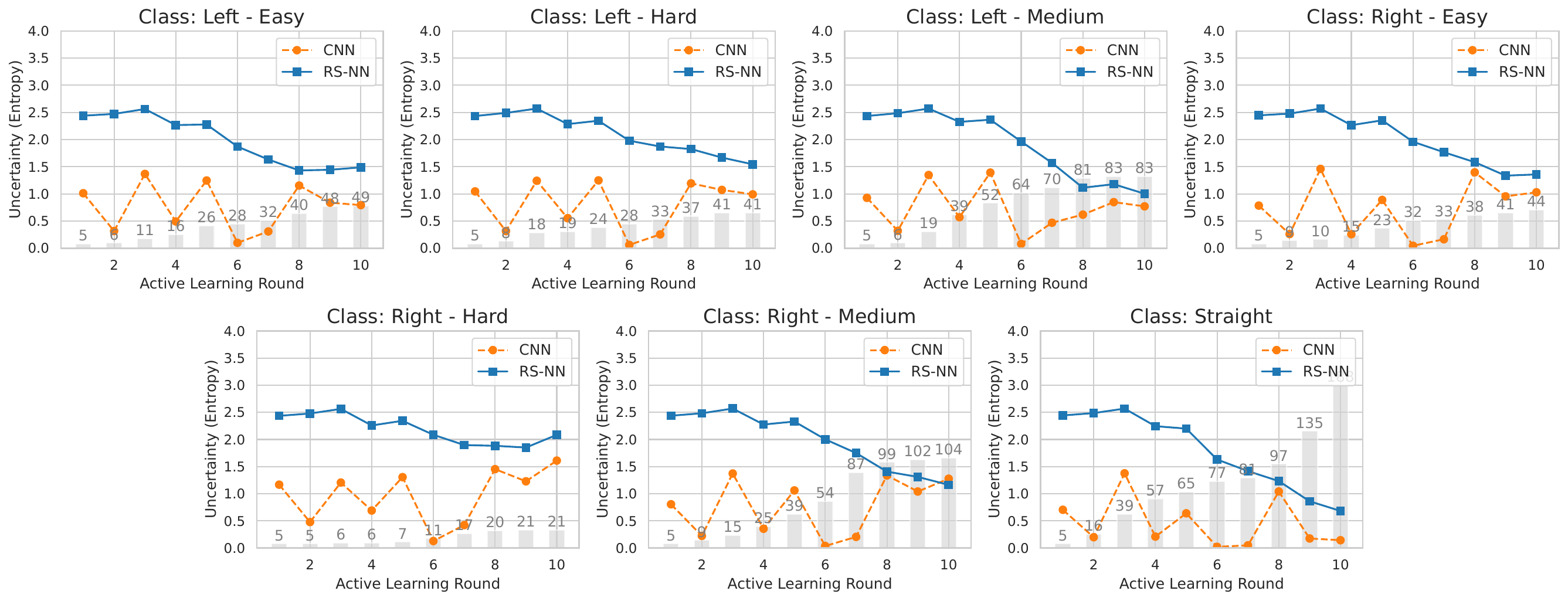}
        \caption{\textit{Experiment 3:} Per-class predictive uncertainty (entropy) across active-learning rounds. RS-NN (blue) begins with higher, honest uncertainty that steadily decreases, whereas CNN (orange) is overconfident early and only learns realistic uncertainty late in training.}
          \vspace{-15pt}
        \label{fig:AL-3-unc}
\end{figure}

\subsubsection{Experiment 3: Fixed Cumulative Class Targets}\label{sec:AL_3}
This experiment enforces strict cumulative targets for labelled samples per class. Training begins with an equal number of examples (e.g., five per class) and increases uniformly each round (10, 15, then 20 per class), while acquisitions still prioritise uncertain instances. This fully balanced scheme demonstrates the upper bound of active learning performance, though it is less practical in real-world data-limited settings. Fig.~\ref{fig:AL-3} shows overall test accuracy under this protocol. With all classes represented from the outset, CNN starts near 40\% once it has at least one sample per class, while RS-NN begins lower ($\approx$15\%) due to its cautious uncertainty estimates but quickly catches up. By round~4 (20 samples/class, total 140), RS-NN reaches $\approx$75\%, surpassing CNN by 5--10 points and plateauing near 78\%. Class-level results (Fig.~\ref{fig:AL-3-acc}) confirm steady improvements across all seven classes, including rare ones like \textit{Right-Hard} and \textit{Left-Hard}. \textit{Straight} achieves near-perfect accuracy by round~3, while subtler turn classes (e.g., \textit{Left-Easy}, \textit{Left-Medium}, \textit{Right-Medium}) exceed 80\% only in later rounds. Grey bars verify that equal numbers of labels are added per class each round. Uncertainty analysis (Fig.~\ref{fig:AL-3-unc}) highlights calibration differences: RS-NN begins with higher entropies ($\approx$2.5 nats) but reduces them consistently below 1.5 by round~6. CNN, by contrast, shows unrealistically low entropy ($<$0.5) early on, reflecting overconfidence, and only adjusts upward when harder examples appear. This again underscores RS-NN’s advantage in both learning speed and calibrated uncertainty under active learning.

\section{CONCLUSIONS}  \label{sec:conclusion}
This work evaluated uncertainty-aware models on an autonomous racing task. RS-NN consistently outperformed both a standard CNN and a Laplace-Bridge Bayesian network in accuracy and showed markedly superior uncertainty calibration. RS-NN’s belief outputs and entropy estimates enabled more data-efficient active learning, rapidly improving minority-class performance under stratified and balanced sampling. Crucially, we validated RS-NN in a live AV control loop: by publishing real-time entropy via ROS, a multi-tier throttle-scaling policy directly translated predictive uncertainty into safer driving behaviour, by modulating speed according to entropy. This mechanism enforced caution in ambiguous scenarios (e.g., fallen cones, random layouts, poor visibility) while maintaining confidence-driven speed when conditions allowed. Thus, perceptual uncertainty was not only quantified but actively informed safer control decisions. Future work will extend this framework to uncertainty aware object detectors with multi-modal fusion (camera, LiDAR, radar). This work has demonstrated the potential of embedding uncertainty-aware perception into AV control pipelines, and aims to promote wider adoption of uncertainty-aware architectures - to usher in a new breed of safer, uncertainty-aware  Autonomous Vehicles.

\bibliographystyle{IEEEtran}
\bibliography{IEEEabrv,biblio}




\addtolength{\textheight}{-12cm}   

\end{document}